\pdfoutput=1
\documentclass[11pt, a4paper, logo]{hzpaper}

\usepackage[authoryear, sort&compress, round]{natbib}
\usepackage{microtype}
\usepackage{hyperref}
\usepackage{url}
\usepackage{booktabs}

\usepackage{lineno}
\usepackage{graphicx}
\usepackage{tcolorbox}
\usepackage{tikz}
\usetikzlibrary{positioning, fit, arrows.meta, shapes.geometric}
\usepackage{wrapfig}
\usepackage{colortbl} 
\usepackage[T1]{fontenc}

\usepackage{xcolor}
\usepackage{mdframed}
\usepackage{fancyvrb}
\usepackage{etoolbox}
\usepackage{changepage}

\usepackage{tikz}
\usepackage{multirow}
\usepackage{graphicx}
\usepackage{nicematrix} 
\usepackage{rotating}
\usepackage{xcolor}
\usepackage{subfiles} 
\usetikzlibrary{decorations.pathreplacing}

\usepackage{algorithm}
\usepackage{algpseudocode}

\definecolor{accentcolor}{HTML}{175E54}
\hypersetup{
	colorlinks=true,
	citecolor=accentcolor,
	linkcolor=accentcolor,
	urlcolor=accentcolor
}

\usepackage{amsmath, amsfonts, bm}

\def\eqref#1{equation~\ref{#1}}

  \def\1{\bm{1}}

\DeclareMathAlphabet{\mathsfit}{\encodingdefault}{\sfdefault}{m}{sl}
\SetMathAlphabet{\mathsfit}{bold}{\encodingdefault}{\sfdefault}{bx}{n}

\usepackage{array}
\newcolumntype{C}[1]{>{\centering\arraybackslash}p{#1}}

\title{AutoLibra: Agent Metric Induction from Open-Ended Human Feedback}

\author{Hao Zhu\raisebox{0.5em}{\includegraphics[height=1em]{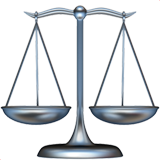}} Phil
Cuvin\raisebox{0.5em}{\includegraphics[height=1em]{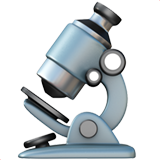}} Xinkai
Yu\raisebox{0.5em}{\includegraphics[height=1em]{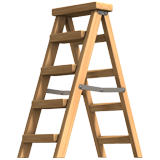}} Charlotte Ka Yee
Yan\raisebox{0.5em}{\includegraphics[height=1em]{figs/scale.png}} Jason Zhang\raisebox{0.5em}{\includegraphics[
	height=1em
]{figs/scale.png}} Diyi Yang\raisebox{0.5em}{\includegraphics[height=1em]{
	figs/scale.png
}}\\ \raisebox{0.5em}{\includegraphics[height=1em]{figs/scale.png}}Stanford
University \raisebox{0.5em}{\includegraphics[height=1em]{figs/microscope.png}}University
of Toronto \raisebox{0.5em}{\includegraphics[height=1em]{figs/ladder.png}}University
of Pennsylvania\\ \texttt{\{zhuhao,ckyy,jasonbz,diyiy\}@stanford.edu}\\ \texttt{philippe.cuvin@mail.utoronto.ca},
\texttt{xinkaiyu@sas.upenn.edu}\\ \href{https://github.com/Open-Social-World/autolibra}{Code}\quad\href{https://huggingface.co/datasets/open-social-world/autolibra}{Data}\quad
Website: \url{https://autolibra.opensocial.world}}
\def\year{2025} 

\providecommand{\theabstract}{}

\begin{abstract}
	Agents are predominantly evaluated and optimized via task success metrics,
	which are coarse, rely on manual design from experts, and fail to reward
	intermediate emergent behaviors. We propose \emph{AutoLibra} \protect
	\includegraphics[height=1em]{figs/scale.png}
	, a framework for agent evaluation, that transforms open-ended human feedback \emph{e.g.}
	``\textsl{If you find that the button is disabled, don't click it again}'', or
	``\textsl{This agent has too much autonomy to decide what to do on its own}'' into
	metrics for evaluating fine-grained behaviors in agent trajectories. AutoLibra
	accomplishes this by grounding feedback to an agent's behavior, clustering similar
	positive and negative behaviors, and creating concrete metrics with clear
	definitions and concrete examples, which can be used for prompting LLM-as-a-Judge
	as evaluators. We further propose two \emph{meta-metrics} to evaluate the alignment
	of a set of (induced) metrics with open feedback: ``coverage'' and ``redundancy''.
	Through optimizing these meta-metrics, we experimentally demonstrate AutoLibra's
	ability to induce more concrete \textbf{agent evaluation} metrics than the ones
	proposed in previous agent evaluation benchmarks and discover new metrics to analyze
	agents. We also present two applications of AutoLibra in \textbf{agent
	improvement}: First, we show that AutoLibra serve human prompt engineers for diagonalize
	agent failures and improve prompts iterative. Moreover, we find that AutoLibra
	can induce metrics for automatic optimization for agents, which makes agents
	improve through self-regulation. Our results suggest that AutoLibra is a powerful
	task-agnostic tool for evaluating and improving language agents.
\end{abstract}
\begin{document}
	\maketitle
	\tableofcontents

\section{Introduction}

Humans readily acquire skills from open-ended instructions and feedback from others
\citep{tomasello1993cultural}. These instructions and feedback are internalized for
self-regulated learning \citep{pintrich2002development,nicol2006formative}, providing
internal signals for continuous improvement. Drawing inspiration from this
process, we investigate how well AI agents can benefit from open-ended human feedback
through induction of generalizable metrics.


In this paper, we introduce AutoLibra \protect
\includegraphics[height=1em]{figs/scale.png}
, a metric induction method, as a novel agent evaluation framework that
mitigates the limitations of current evaluation paradigms. AutoLibra is an evaluation
tool that induces interpretable metrics for AI agents from open-ended human feedback,
which can be collected from end users of AI agents or experts. This offers two advantages:
(1) It is much easier to provide concrete feedback for trajectories than
creating metrics, and (2) AutoLibra allows us to evaluate agents from the perspective
of the users. AutoLibra-induced metrics provide concrete definitions of
behaviors that the model-based evaluation method should look for, which could be
used to understand agent behavior, as well as optimization targets to improve
agents.

\begin{figure}[p]
	\centering
	\includegraphics[width=0.91\linewidth]{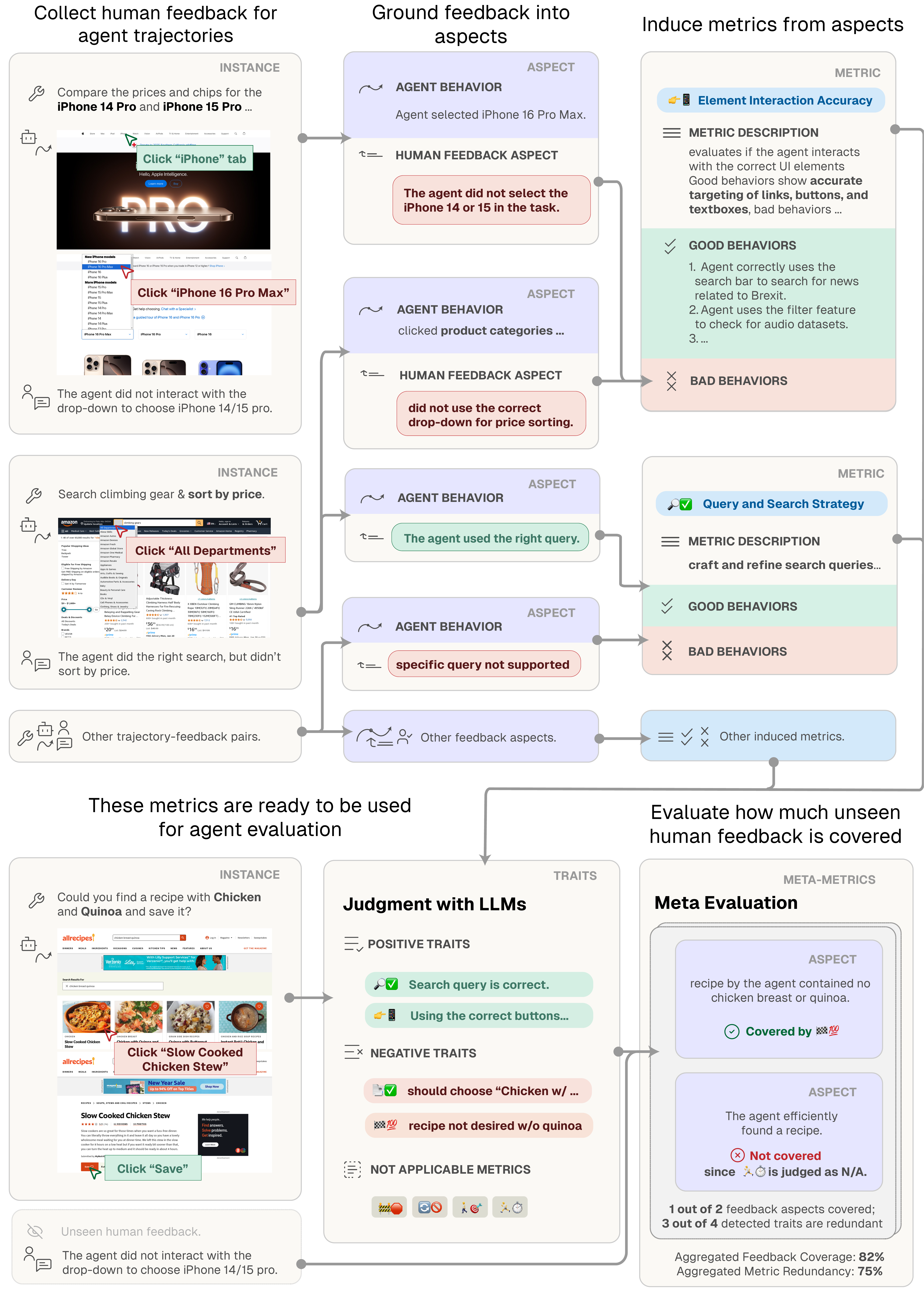}
	\caption{AutoLibra \protect
	\includegraphics[height=1em]{figs/scale.png}
	induces agent evaluation metrics from human feedback, and uses these metrics to
	evaluate agents, which can be meta-evaluated via evaluating the coverage on
	unseen human feedback. Here we show real examples of agent trajectories, human
	feedback, aspects, induced metrics, evaluation results on WebVoyager \citep{he2024webvoyager}.
	\vspace{-20pt}
	}
	\label{fig:teaser}
\end{figure}

Inspired by the code-theme steps of thematic analysis conducted by experts in
social sciences \citep{braun2006using}, we design the AutoLibra induction
process (\S\ref{sec:induction_process}) as two steps: (1) \emph{feedback
grounding}: where we ground every aspect of human feedback on some behavior in
the entire agent trajectory, and (2) \emph{behavior clustering}: where we cluster
the aspects into multiple clusters of similar behaviors to summarize into metrics.
As illustrated in Fig. \ref{fig:teaser}, the user gives a web agent feedback ``the
agent did not choose iPhone 14/15'' which is grounded to the agent's behavior, choosing
``iPhone 16 Pro'' from the drop-down menu. Similar behaviors are clustered into
a common cluster, summarized as \textit{Element Interaction Accuracy}.

The AutoLibra evaluation process is designed to provide a closed-loop feedback
signal for the induction process. The agent trajectories used in the induction process
are scored by LLM-as-a-Judge \citep{zheng2023judging} on the induced metrics. The
evaluation process (\S\ref{sec:evaluation_process}) then tries to match the
feedback aspects, \emph{e.g.} ``recipe does not contain quinoa'', with the traits,
\emph{e.g.} \texttt{task-requirement-\\achievement}. In this way, we can meta-evaluate
the quality of the metrics: (i) \emph{coverage} (what proportion of feedback aspects
can be matched with an agent trait), and (ii) \emph{redundancy} of the metrics (what
proportion of the detected traits are not mentioned by humans). These two metrics
provide an overall statistical picture of the quality of the induced metrics.
Based on these two metrics, we can search for the set of metrics with the lowest
redundancy and the highest coverage. As shown in \S\ref{sec:metric-optimization},
we find that as the number of metrics increases, the redundancy increases, and
the coverage ultimately converges to the maximum coverage. With AutoLibra, our
aim is to answer the following research questions:
\vspace{-8pt}
\begin{itemize}
	\setlength{\itemsep}{0em}

	\item[\textbf{RQ1:}] How well do AutoLibra's step-wise results align with human
		judgment?

	\item[\textbf{RQ2:}] Does AutoLibra provide insights into agent behavior beyond
		expert-designed metrics?

	\item[\textbf{RQ3:}] Can AutoLibra provide optimization signals for
		improving agents' performance?
\end{itemize}
\vspace{-8pt}

\noindent Experiments within multiple agent domains, including collaborative agents \citep{shao2024collaborative},
social agents \citep{zhousotopia}, web agents \citep{zhouwebarena,he2024webvoyager},
and text game agents \citep{paglieri2024balrog,cloos2024babaaibreakrules}, demonstrate
that AutoLibra is able to induce fine-grained and interpretable metrics with high
coverage and low redundancy in unseen human feedback with 80 trajectories per dataset
annotated with one feedback for each. These metrics are more concrete, and some of
them were even overlooked in expert designed metrics or error analysis (\S\ref{sec:lens}).
AutoLibra can iteratively discover new, emergent metrics (\S\ref{sec:iterative-induction})
throughout the agent optimization process, and provide optimization signals
helps improve the performance of frontier LLM in a challenging 2D text game by
over 20\% (\S\ref{sec:ladder}) in 3 stages with only 18 trajectory annotated per
stage.

%

\section{\texorpdfstring{AutoLibra
\includegraphics[height=1em]{figs/scale.png}}{AutoLibra}}

To address the limitations of existing evaluation paradigms, AutoLibra \protect
\includegraphics[height=1em]{figs/scale.png}
is designed to meet the following desiderata: (1) \emph{induced from agent
behavior}: This ensures that metrics are grounded in agent trajectories rather than
predefined by human experts, (2) \emph{self-validating}: Allows choosing minimal
set of metrics that cover unseen human feedback with sufficient abstraction to
be useful across different tasks, and (3) \emph{generalizable}: Applicable to various
agent environments, independent of domain-specific design. Based on feedback data
collected from humans (\S\ref{sec:collecting-human-feedback}), AutoLibra achieves
these desiderata through a closed-loop pipeline consisting of two processes:
\textbf{Induction Process} that converts agent behaviors and corresponding feedback
into metrics, (\S\ref{sec:induction_process}) and \textbf{Evaluation Process}
that predicts ratings and quality of new agent behaviors on the induced metrics
(\S\ref{sec:evaluation_process}).

\subsection{Collecting human feedback}
\label{sec:collecting-human-feedback} In this paper, we use human feedback from
two groups: (1) End-users -- for agents that interact directly with humans, we use
the feedback from the users who interact and converse with the agents. CoGym
\citep{shao2024collaborative} is the environment that belongs to this category,
and we use the user comments collected in their study, resulting in 197 trajectories
with feedback. (2) Experts -- for agents that do not directly interact with humans,
we use the feedback from human annotators (five authors in this paper) who
observe agent trajectories. All other environments belong to this category,
these being Sotopia \citep{zhousotopia}, WebArena \citep{zhouwebarena},
WebVoyager \citep{he2024webvoyager}, Baba-is-ai \citep{cloos2024babaaibreakrules},
and MiniHack \citep{samvelyan2021minihackplanetsandboxopenended}. For each
trajectory, we collect only one element of feedback based on the complete agent trajectories.\footnote{While
in theory we can leverage feedback on specific steps to achieve better feedback
grounding and multiple feedback for single trajectory, we leave it as future work.}

Annotators are instructed to explicitly indicate the aspects of agent behavior
that they classify as good or bad, and to avoid general comments such as \textsl{"The
agent is good at solving the task"}. The annotators can also choose from a terminal
or a web interface; in both cases the annotator is provided with the agent's
task and then view the agent's observation and actions step by step, in text form.\footnote{While
viewing screenshots is standard for web navigation tasks, we keep the observation
format consistent across agents and humans to encourage more grounded feedback.}
For multi-agent tasks, we annotate each agent's trajectory in a given
interaction separately. For Sotopia \citep{zhousotopia}, WebArena \citep{zhouwebarena},
and WebVoyager \citep{he2024webvoyager}, we annotate 100 trajectories of agents based
on GPT-4 \citep{achiam2023gpt} with feedback for each dataset. For experiments in
\S\ref{sec:ladder} we annotate 18 trajectories for each dataset in each
iteration. The annotation process is fast: Human annotators spend less than 5
minutes to provide feedback for each trajectory; \S\ref{sec:lens}, we randomly hold
out 20\% of the trajectories for validation.


\subsection{Induction Process}
\label{sec:induction_process} \textbf{Feedback Grounding} The feedback of human
annotators can contain multiple aspects; e.g. \textsl{``AI agent was pretty good
at giving me a consistent itinerary and vacation plan, although it froze on the last
couple of minutes.''}, collected from human annotators in CoGym \citep{shao2024collaborative},
contains a positive aspect about the agent's ability to generate a consistent
itinerary, and a negative aspect about the agent freezing at the end. Here we
define an \emph{aspect} as a triple $(\texttt{behavior}, \texttt{feedback}, \texttt
{sign})$. In the positive aspect of the previous example, the \texttt{behavior}
is the agent's actions to create a 20-day itinerary for the Maldives, the \texttt{feedback}
is that the created itinerary is consistent and the \texttt{sign} is positive.
This grounding procedure is similar to the coding procedure in thematic analysis.

We feed the trajectory and the feedback into the LLM (we use GPT-4o \citep{openai2024gpt4ocard}
as it yields good results in our pilot experiments) and prompt the LLM with the following
instructions: (1) break down the feedback into bullet points; (2) for each bullet
point, find the corresponding part of the trajectory to which the feedback
refers. Finally, we use constrained decoding to force GPT-4o to output the
aspects in the previous format. In our experiments, we find that on most
datasets, for each trajectory, the LLM can generate one to five aspects, with a mean
of one to two aspects.

\paragraph{Behavior Clustering}

The second step of the extraction process is to group the aspects into $N$ metrics.
To illustrate this step, we consider another example in the same dataset \textsl{``The
AI responds quickly to write and run the Python script``} where the \texttt{behavior}
is the agent's action to quickly write and run a Python script, the \texttt{feedback}
is that the agent responds quickly, and the \texttt{sign} is positive. Although
this aspect is a positive aspect, it reflects the same dimension of the agent's behavior
as the previous negative aspect, with an opposite value. Each \emph{metric} is a
cluster of aspects, with a definition summarizing the criteria of positive behaviors,
a list of positive behavior examples, and a list of negative behavior examples.
This clustering procedure is similar to the theme induction step in thematic
analysis.

However, clustering similar agent behaviors together is challenging for statistical
clustering methods.\footnote{ In preliminary experiments, we tried to use K-means
clustering on the aspect vectors generated by embedding model \texttt{text-embedding-3-large},
but the clusters are mostly based on tasks and not on the behaviors. } Inspired by
LLM-based semantic clustering and concept induction methods \citep{viswanathan2024large,lam2024concept},
we prompt an LLM (o3-mini high\footnote{https://openai.com/index/openai-o3-mini/},
as it produces the most accurate coverage and redundancy scores as evaluated
later) to cluster the aspects into metrics. As illustrated in Fig. \ref{fig:behavior_clustering},
we gather all the aspects of $M$ trajectories and cluster into $N$ metrics, where
$N$ is a parameter set through the optimization process (\S\ref{sec:metric-optimization}).
We provide the LLM with the following instructions: \emph{The granularity of the
grouping should be minimal; only very similar behaviors are grouped together;
but don't limit to one particular website or one particular character}, which
empirically makes the metrics more concrete but still applicable across
different tasks.

\subsection{Evaluation Process}
\label{sec:evaluation_process}

\paragraph{Evaluating agents with induced metrics}

LLM-as-a-Judge \citep{zheng2023judging}, or more broadly, model-based evaluation
\citep{zhang2019bertscore,celikyilmaz2021evaluationtextgenerationsurvey} is a method
to use machine learning models to evaluate the output of other machine learning models.
The success of LLM-as-a-Judge depends on the gap between the difficulty of
evaluation or verification and that of generation and action. In agentic tasks,
this gap is often large, as the policy model must perform multiple steps in decision-making,
while the evaluation model must only classify the trajectories, which make LLM-as-a-Judge
widely used \citep{zhouwebarena,he2024webvoyager,zhousotopia}. In AutoLibra, we
employ LLM-as-a-Judge to evaluate the agent trajectories configured with the
induced metrics. However, LLM-as-a-Judge can be replaced by any other evaluation
methods implementing the induced metrics; \emph{e.g.} an \texttt{interact-valid-element}
metric could be evaluated by a rule-based evaluator that checks if the agent interacts
with valid elements on the webpage. Wenote that AutoLibra could be used with
other evaluation methods, such as programmatic evaluation \citep{maeureka}; we leave
generating programs for the induced metrics for future work.

As illustrated in Fig. \ref{fig:llm_as_a_judge}, taking the induced metrics as input,
an LLM (we use o3-mini medium, as it provides similar results in this step to o3-mini
high) is prompted to rate the agent trajectories to \{+ 1, -1, N/A\} for each
metric. For an agent trajectory, the metrics labeled +1 are the positive \emph{traits},
and the ones labeled -1 are the negative \emph{traits}. When we calculate the
scores of the metrics, we use the ratio of agent trajectories rated as positive to
the ones that are rated as positive or negative, ignoring those rated as N/A, since
not all metrics are applicable to all trajectories (some metrics like \texttt{valid-search-terms}
are only applicable when the task involves searching).

\paragraph{Meta evaluation}
The final loop component is the meta-evaluation, i.e. evaluating the evaluation metrics
induced by AutoLibra. This step matches the traits detected by the LLM-as-a-Judge
with aspects grounded from the human feedback. The goal is to verify whether (1)
the induced metrics cover the behaviors the human annotators care about, and (2)
LLM-as-a-Judge can produce accurate evaluation results based on the induced
metrics. In the previous example, if the \texttt{respond-promptly} is extracted as
a metric, and the LLM-as-a-Judge has the same opinion as the human annotators, then
this aspect is considered as successfully covered. If either a similar metric
was not extracted, or the LLM-as-a-Judge assigns a different score, then this
aspect is considered as not covered.

\begin{figure}[t]
	\centering
	\includegraphics[width=0.8\linewidth]{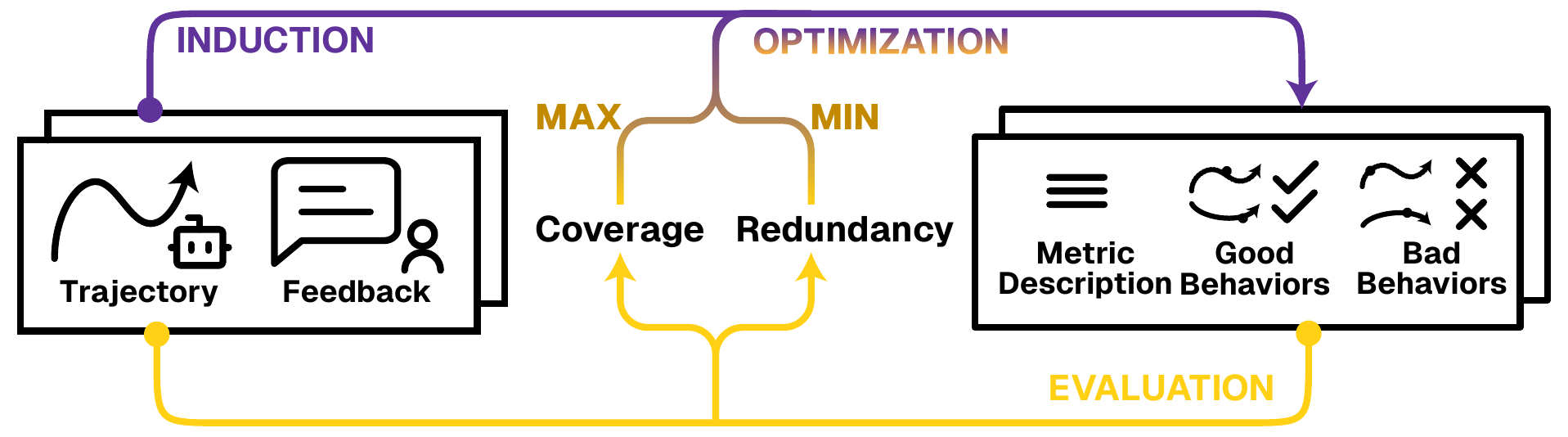}
	\caption{Metric optimization: optimizing the induction process through maximizing
	the coverage while minimizing redundancy of the metrics, calculated via the
	evaluation process.}
	\label{fig:autolibra_optimization}
\end{figure}

As illustrated in Fig. \ref{fig:meta_evaluation}, we perform meta-evaluation for
each trajectory-feedback pair by classifying the aspects into positive and negative
aspects, classifying traits into positive and negative traits based on rating, then
matching the positive aspects with positive traits and the negative aspects with
negative traits. We prompt an LLM (we use GPT-4o \citep{openai2024gpt4ocard}) with
a list of aspects and another list of traits and ask the LLM to find the best matching
trait for each aspect or decide that there is no matching trait. The \emph{coverage}
of the whole dataset is calculated as the proportion of aspects of all instances
that have a matching trait, and the \emph{redundancy} is calculated as the proportion
of traits of all instances that have not been matched with any aspect.

\section{Optimizing and Validating AutoLibra \protect
\includegraphics[height=1em]{figs/scale.png}}
AutoLibra is designed to be self-validating through the evaluation process,
which allows us to search the optimal set of metrics that cover the human
opinion the best (\S\ref{sec:metric-optimization}). This optimization process
can also be applied iteratively throughout the agent improvement process. As the
agent is optimized, new metrics can be added to existing metrics (\S\ref{sec:iterative-induction}),
which is similar to how unit tests are kept throughout software development to
prevent new features from interfere with existing features. In the last part of this
section, we study the alignment between each step of AutoLibra and human
judgment.

\subsection{Metric Optimization}
\label{sec:metric-optimization}

Illustrated in Fig. \ref{fig:autolibra_optimization}, we optimize the metric induction
process to maximize \emph{coverage} and minimize \emph{redundancy}. Among the
two, we prioritize coverage of the metrics to provide a comprehensive evaluation
of the agent behavior, while minimizing overlap within the metrics to avoid
redundancy, thus maximizing the utility of induced metrics. To optimize for this
objective, we generate 20 different sets of metrics, with metric count $N$
ranging from 4 to 13, and calculate the coverage and redundancy of the metrics in
human feedback. We then select metrics with a coverage of at least the highest
coverage minus 1\%, and the lowest redundancy. This is performed iteratively, by
resetting the range of $N$ to the number of metrics selected previously $\pm$2,
repeating until the coverage and redundancy of the selected metrics converge, normally
within 3 iterations. While this optimization process is simple, experiments with
various other
\begin{figure}[!t]
	\centering
	\includegraphics[width=0.5\textwidth]{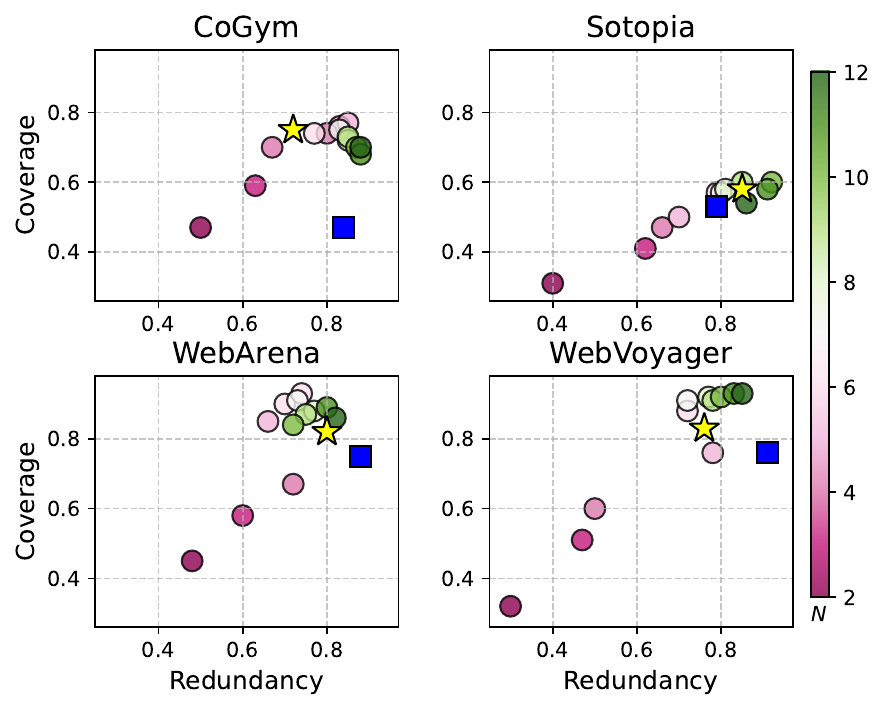}
	\caption{Coverage and redundancy of AutoLibra metrics on four agentic datasets.
	Circles indicate coverage and redundancy for different induced metrics; stars
	indicate the the best metrics' coverage and redundancy on held-out human feedback;
	squares show an ablation test, indicating the effect when good and bad behavior
	examples are removed from metrics, demonstrating the criticality of concrete
	behavior examples}
	\label{fig:coverage-redundancy}
\end{figure}
optimization strategies, including genetic algorithms and iterative
clustering saw none of them yield better results than the simple strategy. Fig. \ref{fig:coverage-redundancy}
shows the highest coverages of the metrics of size $N$, which converge around
$N=6$ to $10$ depending on the datasets. The best coverage on Sotopia \citep{zhousotopia}
is the lowest among all four datasets, $60\%$, likely due to the diversity of
the tasks in the dataset, while coverage on WebArena \citep{zhouwebarena} and
WebVoyager \citep{he2024webvoyager} are the highest, $88\%$. We also find that the
coverage of the held-out trajectories is only slightly worse ($<5\%$) than the trajectories
we use to induce the metrics, which is expected since we use the exact examples
extracted from the latter. Lastly, we show that the good and bad behaviors are
crucial in the metrics, dropping which resulting in up to $30\%$ coverage decrease
on CoGym. 

\subsection{Iterative Metric Induction}
\label{sec:iterative-induction} When applying AutoLibra to agent optimization, we
can iteratively induce new metrics, as agents develop new failure modes or new
behaviors as they improve, which is useful for tracking agents' progress across
different iterations.\footnote{Alternatively, a new set of metrics can be induced
from scratch for each iteration - in practice, we do not find that this results
in any coverage loss, but we choose the former method for consistency} To do
this, we modify the behavior clustering step: we provide the LLM with the
existing metrics and definitions, asking the LLM not to change the definitions
of the existing metrics, to only add new behaviors to the existing metrics, and to add
new metrics if necessary. We apply the same optimization strategy as in the
metric optimization step ensure the newly induced metrics cover emerging
behaviors and do not overlap with existing metrics.
\smallskip

\subsection{How aligned are the steps in AutoLibra with human judgment?}
Since AutoLibra uses LLMs in each step, we first ask whether LLM outputs are
reliable or aligned with human judgment. To measure the alignment of AutoLibra metric
induction with human judgment, we validate the feedback grounding, agent evaluation,
and meta evaluation steps by having human experts manually review each step (with
exception of the behavior clustering step, as it is prohibitively time-intensive
for human annotators to process and cluster more than 400 aspects), scoring (1/0)
based on whether they agree with the outcomes of each iteration. The coverage and
redundancy scores, in combination with the validation results of the other steps
in the loop, thus serve as an indirect validation for the behavior clustering step.
Table \ref{tab:validation} shows the agreement rate of human annotators in
AutoLibra steps. It should be noted that these tasks are significantly different;
e.g., grounding for WebVoyager \citep{he2024webvoyager} is challenging due to the
length and wide action space of the trajectory, and LLM-as-a-Judge for Sotopia
\citep{zhousotopia} is difficult due to the complexity of the evaluation of social
interactions. Our results show that the majority (significantly over 85\%) of results
in AutoLibra are reliable according to human validation.

\begin{table}[!h]
	\centering
	\small
	\caption{ The ratio of instances marked as fully correct in human validation.
	For each step and each task, we randomly sample 40 instances to reach a relatively
	small confidence interval of $0.04$ and ask human annotators to label them as completely
	correct or not. Although the agreement scores vary across tasks and steps, the
	average agreement for each step and dataset is above 0.85 significantly. }
	\begin{tabular}{cccccc|c}
		\toprule Steps     & CoGym & Sotopia & WebArena & WebVoyager & Baba-is-AI & Average           \\
		\midrule Grounding & 0.95  & 0.95    & 0.98     & 0.93       & 0.93       & 0.95 ($\pm 0.03$) \\
		LLM-as-a-Judge     & 0.90  & 0.85    & 0.95     & 1.00       & 0.90       & 0.92 ($\pm 0.04$) \\
		Meta-Evaluation    & 0.98  & 0.90    & 0.85     & 0.83       & 0.95       & 0.90 ($\pm 0.04$) \\
		\bottomrule
	\end{tabular}

	\label{tab:validation}
\end{table}


    \section{\texorpdfstring{AutoLibra as A Lens
    \includegraphics[height=1em]{figs/microscope.png}
    : Agent Evaluation with AutoLibra}{AutoLibra as a lens: agent evaluation with
    AutoLibra}}
    \label{sec:lens}

    In this section, we use AutoLibra as a lens to provide grounded, behavior-salient
    insights into agent trajectories. In three data sets, CoGym \citep{shao2024collaborative},
    Sotopia \citep{zhousotopia}, and WebVoyager \citep{he2024webvoyager}, we
    compare induced metrics with heuristically proposed evaluation dimensions
    and failure modes summarized by the authors. We find that AutoLibra can discover
    more concrete metrics than heuristically defined categories, and novel
    metrics that are overlooked by experts. Tab. \ref{tab:merged_comparison}
    summarizes the comparison between AutoLibra-induced metrics and evaluation
    criteria across the three aforementioned datasets.

    \paragraph{CoGym}
    For CoGym \citep{shao2024collaborative}, AutoLibra induces 9 metrics from feedback
    from \textbf{end users}, which can correspond to the five failure categories
    proposed by the authors. The failure rate (frequency of a metric score of -1)
    measured by AutoLibra also roughly matches the failure rate of the manually labeled
    CoGym categories by the authors. This shows that AutoLibra induces metrics
    that reflect human-expert categorization and provide an automated
    measurement of agent failures.
    \definecolor{comm}{RGB}{230,242,255}
    \definecolor{sit}{RGB}{255,240,230}
    \definecolor{plan}{RGB}{230,255,230}
    \definecolor{env}{RGB}{255,230,255}
    \definecolor{pers}{RGB}{255,255,230}
    \definecolor{goal}{RGB}{242,242,255}
    \definecolor{believ}{RGB}{255,242,242}
    \definecolor{navstuck}{RGB}{242,255,242}
    \definecolor{hall}{RGB}{255,242,255}
    \definecolor{misalign}{RGB}{245,245,230}
    \definecolor{unmatched}{RGB}{240,240,240}

    \begin{table}[p]
        \centering
        \renewcommand{\arraystretch}{1.2} 
        \small
        \caption{AutoLibra-induced metrics and expert-proposed evaluation
        dimensions and failure categories. (Percentage \%) denotes failure frequency
        or score from AutoLibra or the original papers.}
        \begin{tabular}{@{}lp{\dimexpr0.50\textwidth}p{\dimexpr0.40\textwidth}@{}}
            \toprule                                                                                          & \textbf{AutoLibra\protect\includegraphics[height=1em]{figs/scale.png}-induced metrics}                                                                                                                                                        & \textbf{Failure categories by experts}                                    \\
            \midrule \multirow{11}{*}{\rotatebox[origin=c]{90}{\textbf{CoGym} \citep{shao2024collaborative}}} & \multicolumn{2}{c}{Matched metrics and failure categories}                                                                                                                                                                                     \\
            \cmidrule(lr){2-3}                                                                                & \cellcolor{comm}\textit{Responsiveness and Efficiency} (75\%)                                                                                                                                                                                 & \cellcolor{comm}                                                          \\
                                                                                                              & \cellcolor{comm}\textit{Communication Clarity \& Notification} (8\%)                                                                                                                                                                          & \multirow{-2}{*}{\cellcolor{comm}\textit{Communication} (65\%)}           \\
                                                                                                              & \cellcolor{sit}\textit{Instruction Adherence \& Follow-Through} (24\%)                                                                                                                                                                        & \cellcolor{sit}\textit{Situational Awareness} (40\%)                      \\
                                                                                                              & \cellcolor{plan}\textit{Iterative Refinement and Adaptability} (47\%)                                                                                                                                                                         & \cellcolor{plan}                                                          \\
                                                                                                              & \cellcolor{plan}\textit{Autonomy and Proactiveness} (28\%)                                                                                                                                                                                    & \multirow{-2}{*}{\cellcolor{plan}\textit{Planning} (39\%)}                \\
                                                                                                              & \cellcolor{env}\textit{Content Quality and Coherence} (16\%)                                                                                                                                                                                  & \cellcolor{env}                                                           \\
                                                                                                              & \cellcolor{env}\textit{Search and Retrieval Accuracy} (13\%)                                                                                                                                                                                  & \cellcolor{env}                                                           \\
                                                                                                              & \cellcolor{env}\textit{Data Analysis Competence} (2\%)                                                                                                                                                                                        & \multirow{-3}{*}{ \cellcolor{env}\textit{Environmental Awareness} (28\%)} \\
                                                                                                              & \cellcolor{pers}\textit{Interface and User Experience} (23\%)                                                                                                                                                                                 & \cellcolor{pers}\textit{Personalization} (16\%)                           \\
            \midrule \multirow{13}{*}{\rotatebox[origin=c]{90}{\textbf{Sotopia} \citep{zhousotopia}}}         & \multicolumn{2}{c}{Matched metrics and social dimensions}                                                                                                                                                                                      \\
            \cmidrule(lr){2-3}                                                                                & \cellcolor{goal}\textit{Goal Achievement \& Outcome Effectiveness} (19\%)                                                                                                                                                                     & \cellcolor{goal}\textit{Goal Completion} (14\%)                           \\
                                                                                                              & \cellcolor{believ}\textit{Conversational Naturalness \& Efficiency} (5\%)                                                                                                                                                                     & \cellcolor{believ}                                                        \\
                                                                                                              & \cellcolor{believ}\textit{Personality Consistency and Alignment} (2\%)                                                                                                                                                                        & \cellcolor{believ}                                                        \\
                                                                                                              & \cellcolor{believ}\textit{Contextual Integration of Identity} (1\%)                                                                                                                                                                           & \multirow{-3}{*}{\cellcolor{believ}\textit{Believability} (4\%)}          \\
            \cmidrule(lr){2-3}                                                                                & \multicolumn{2}{c}{Unmatched AutoLibra\protect\includegraphics[height=1em]{figs/scale.png}-induced metrics}                                                                                                                                    \\
            \cmidrule(lr){2-3}                                                                                & \multicolumn{2}{C{0.93\textwidth}}{\cellcolor{unmatched}\textit{Negotiation Tactics and Strategic Adaptability} (14\%), \textit{Responsiveness and Conversational Termination} (5\%), \textit{Adaptability and Flexibility in Dialogue} (7\%)} \\
            \cmidrule(lr){2-3}                                                                                & \multicolumn{2}{c}{Unmatched Sotopia-Eval dimensions}                                                                                                                                                                                          \\
            \cmidrule(lr){2-3}                                                                                & \multicolumn{2}{C{0.93\textwidth}}{\cellcolor{unmatched}\textit{Relationship, Knowledge, Secret, Financial and Material Benefits, Social Rules}}                                                                                               \\
            \midrule \multirow{14}{*}{\rotatebox[origin=c]{90}{\textbf{WebVoyager} \citep{he2024webvoyager}}} & \multicolumn{2}{c}{Matched metrics and failure reasons}                                                                                                                                                                                        \\
            \cmidrule(lr){2-3}                                                                                & \cellcolor{navstuck}\textit{Error Recovery \& Adjustment} (15\%)                                                                                                                                                                              & \cellcolor{navstuck}                                                      \\
                                                                                                              & \cellcolor{navstuck}\textit{Step Efficiency \& Action Redundancy} (13\%)                                                                                                                                                                      & \cellcolor{navstuck}                                                      \\
                                                                                                              & \cellcolor{navstuck}\textit{Navigation Accuracy} (11\%)                                                                                                                                                                                       & \cellcolor{navstuck}                                                      \\
                                                                                                              & \cellcolor{navstuck}\textit{Access Barrier Handling} (2\%)                                                                                                                                                                                    & \multirow{-4}{*}{\cellcolor{navstuck}\textit{Navigation Stuck} (44\%)}    \\
                                                                                                              & \cellcolor{hall}\textit{Information \& Verification Accuracy} (16\%)                                                                                                                                                                          & \cellcolor{hall}\textit{Hallucination} (22\%)                             \\
                                                                                                              & \cellcolor{misalign}\textit{Result Relevance Accuracy} (9\%)                                                                                                                                                                                  & \cellcolor{misalign}\textit{Prompt Misalignment} (9\%)                    \\
            \cmidrule(lr){2-3}                                                                                & \multicolumn{2}{c}{Unmatched AutoLibra\protect\includegraphics[height=1em]{figs/scale.png}-induced metrics}                                                                                                                                    \\
            \cmidrule(lr){2-3}                                                                                & \multicolumn{2}{C{0.93\textwidth}}{\cellcolor{unmatched}\textit{Query and Search Strategy Efficiency} (7\%), \textit{Final Output and Summarization Quality} (18\%)}                                                                           \\
            \cmidrule(lr){2-3}                                                                                & \multicolumn{2}{c}{Unmatched WebVoyager fail reasons}                                                                                                                                                                                          \\
            \cmidrule(lr){2-3}                                                                                & \multicolumn{2}{C{0.93\textwidth}}{\cellcolor{unmatched}\textit{Visual Grounding Issue} (25\%)}                                                                                                                                                \\
            \bottomrule
        \end{tabular}

        \label{tab:merged_comparison}
    \end{table}

    \paragraph{Sotopia}
    Sotopia \citep{zhousotopia} proposed 7 dimensions for evaluating social intelligence
    in AI agents. With AutoLibra, we recover the exact dimension \emph{Goal
    Completion}, and 3 metrics as the subdimensions of \emph{Believability}, indicating
    that \textit{Believability} could be too high-level, while AutoLibra provides
    more concrete breakdown metrics. The failure rate (frequency of a score of -1
    metric rating, indicating the agent performs poorly on that metric) measured
    by AutoLibra in these two categories roughly matches the score of the
    Sotopia dimensions of the agent we studied. AutoLibra induces another four metrics
    overlooked in the heuristically proposed Sotopia-Eval dimensions. We note
    that the other five dimensions in Sotopia are still valuable evaluation
    dimensions for social intelligence. However, behaviors captured by
    dimensions \emph{Financial and Material Benefits}, \emph{Knowledge}, and
    \emph{Secret} are often also captured by \textit{Goal Completion} and
    \textit{Believability}. As a result, AutoLibra produces the single \textit{Goal
    Achievement and Outcome Effectiveness} by minimizing redundancy. Whereas,
    \textit{Relationship} and \textit{Social Rules} captures long-tailed
    behaviors not captured by AutoLibra.

    \paragraph{WebVoyager}
    Similarly, for web navigation tasks, AutoLibra also discovers metrics such as
    \textit{Access Barrier Handling}, \textit{Error Recovery and Adjustment}, \textit{Step
    Efficiency and Action Redundancy}, and \textit{Navigation Accuracy}, which much
    more closely reflect concrete agent behavior than the failure analysis categories
    proposed in previous work \citep{he2024webvoyager,zhou2024proposeragentevaluatorpaeautonomousskilldiscovery},
    where they are often simply classified as ``navigation stuck''. We also find
    additional metrics that are not mentioned in the failure analysis, such as \textit{Query
    and Search Strategy Efficiency} and \textit{Final Output and Summarization
    Quality}, which are frequent issues (with frequencies of 7\% and 18\%). Since
    AutoLibra only observes the behavior of the agents, it cannot interpret the
    neural representation, not able to capture the visual grounding issues,
    which are mentioned in the WebVoyager paper. This further demonstrates AutoLibra's
    utility in extracting behavior-salient metrics, and particularly its ability
    to obtain \textbf{fine-grained metrics} that expert design would not have been
    able to extract.

\section{AutoLibra as A Ladder \protect
\includegraphics[height=1em]{figs/ladder.png}
: Agent Improvement with AutoLibra}
\label{sec:ladder}

\begin{figure}[!b]
	\centering
	\includegraphics[width=0.8\textwidth]{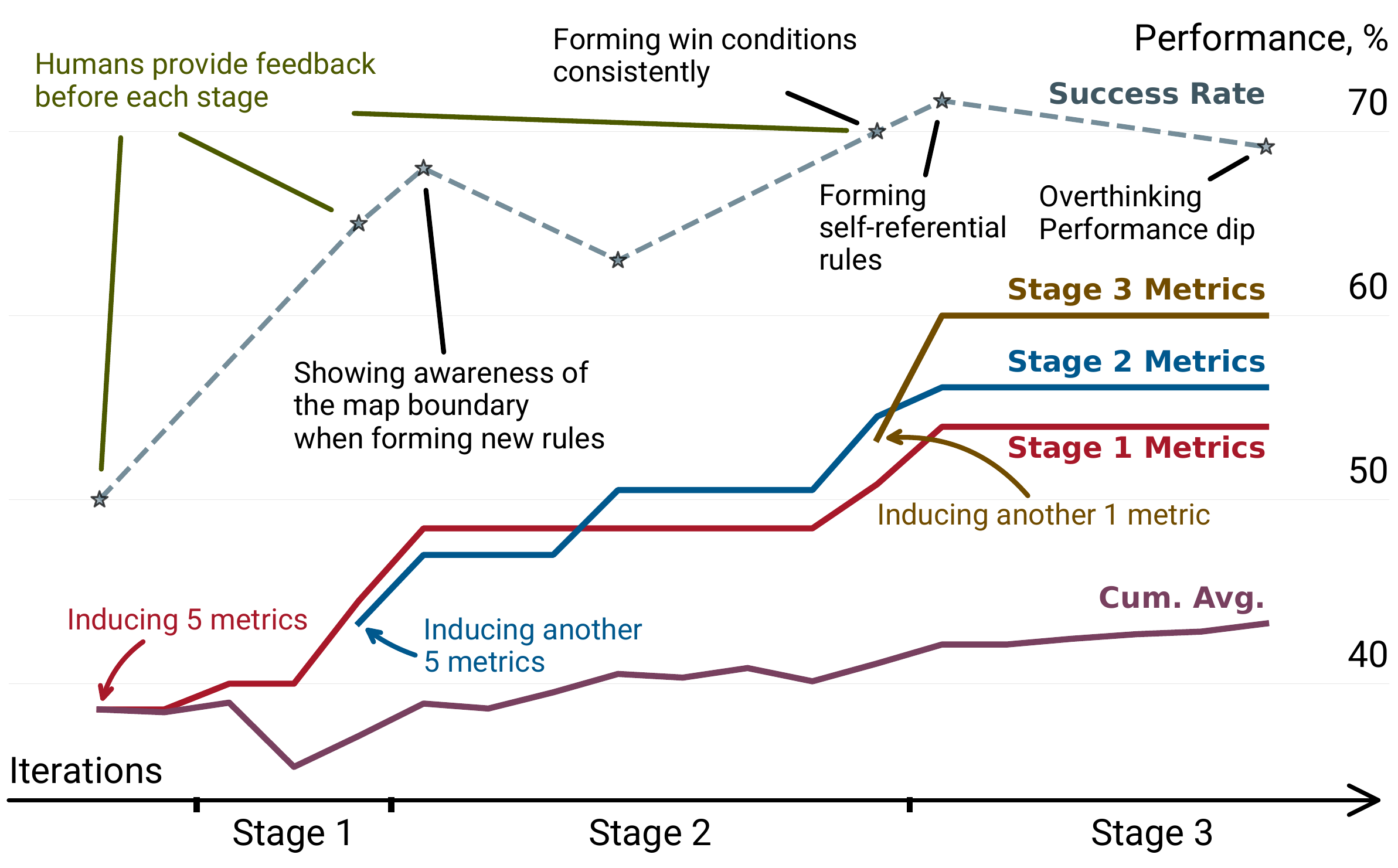}
	\caption{AutoLibra iteratively induce metrics and improves the agent prompts
	through optimizing for the induced metrics. Although not optimized for, the
	success rate of the agent continuously improve until Stage 3, when the agent begins
	to overthink.}
	\label{fig:autolibra_self_improving}
\end{figure}

\begin{figure}[!t]
\centering
\includegraphics[width=0.5\textwidth]{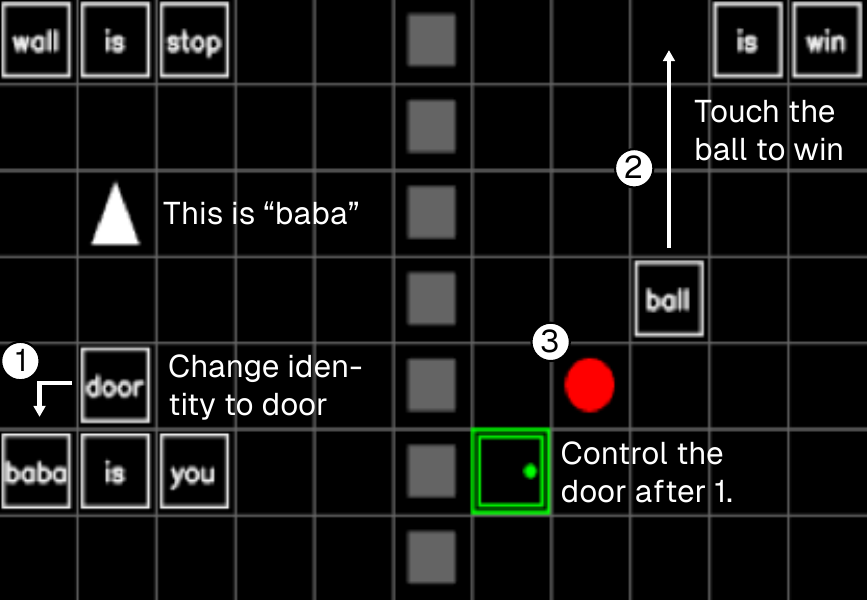}
\caption{Example of Baba-Is-AI game.}
\label{fig:baba-is-ai}
\end{figure}

As AutoLibra can automatically induce metrics from human feedback, a natural
question to ask is whether it can enable self-regulated improvement in agents
through iterative feedback. This can be achieved through optimizing the agent prompts
towards higher scores on the metrics extracted by AutoLibra. To answer this
question, we use a challenging 2D game Baba-Is-AI \citep{cloos2024babaaibreakrules, paglieri2024balrog}
as a benchmark. Inspired by \href{https://hempuli.com/baba/}{Baba-Is-You}, this
game requires not only following rules to achieve goals, but also manipulating the
rules, even self-referential ones. For example, in the game illustrated in Fig.
\ref{fig:baba-is-ai}, the agent needs to change self-referential rules from \textsl{baba
is you}, to \textsl{door is you} to control the green door on the other side of
the wall, form a new win rule \textsl{ball is win}, and navigate to the red ball
to achieve the win condition. To achieve a high score on this dataset, the agent
needs not only planning, but also metacognitive skills, which is very challenging
for LLM agents with frontier models as shown in the Balrog benchmark \citep{paglieri2024balrog}.
In this experiment, we use Gemini-2.5-Flash \citep{geminiteam2025geminifamilyhighlycapable}
for the agent, AutoLibra, and agent prompt optimization, throughout the
experiment, which will be referred as the LLM in this section. Gemini-2.5-Flash
is ranked as the 3rd place, with a success rate of $50.8\%\pm 4.6\%$ on the Balrog
leaderboard for Baba-is-AI at the time of submission, and the state-of-the-art
result is $56.7\%\pm4.5\%$. We chose this model due to the tradeoff between the cost and
the performance. 

Fig. \ref{fig:autolibra_self_improving} illustrated our procedure, and
summarized the results. In this experiment, we employ an iterative process by
improving the agents in 3 stages through providing human feedback on 6 out of 40
tasks in the Baba-Is-AI \citep{paglieri2024balrog} benchmark. In this way, we can study
if the feedback provided for training tasks can be generalized to unseen tasks.

Before each stage we show human annotators 3 trajectories per task for the 6 tasks,
gather the feedback, and apply AutoLibra iterative metric induction process (\S\ref{sec:iterative-induction}).
This results in 5 metrics for Stage 1 and 2, and another 1 metric for Stage 3. Within
each stage, we iteratively feed 1 LLM agent trajectory on each of these 6 tasks,
together with evaluation results based on these AutoLibra-induced metrics to the
LLM to improve the prompt of the LLM agent. This process results in continuous
improvement not only on the running maximum metric scores, the cumulative average
metrics, but also game success rate. Fig. \ref{fig:autolibra_self_improving}
shows these statistics on the whole 40 tasks, although we only use 6 out of the 40
tasks in the whole optimization process. Upon examining the agent trajectories, we
find the skills learned in the process. In the first stage, the agent learns to find
rules to form based on the map boundary, which could be a result of an induced
metric \texttt{map-n-constraint-recognition}. Similarly, more advanced skills are
learned in Stage 2 and 3, including forming win conditions and self-referential
rules, probably as a result of metric \texttt{rule-manipulation-proficiency}.

Our results show that the metrics induced by AutoLibra form effective objectives
for improving the agents through prompt optimization. It should note that AutoLibra
is a metric induction method, which is orthogonal to learning algorithms, including
prompt optimization, fine-tuning or reinforcement learning. We show that this process
improves agent success rate by 20\% without optimizing for success rate, and in the
future, researchers can study the effect of employing other learning algorithm.

\section{Related Work}
AutoLibra unifies three areas of research: it draws inspiration from \textit{thematic
analysis} to create \textit{nautral language-derived evaluation metrics} to evaluate
and reward \textit{AI agents}.

\noindent\textbf{Evaluating AI agents}
Much of the work in AI agent evaluation focuses around benchmarks which contains
both task suites and evaluation metrics. In addition to the datasets we used in
this paper, SWE-Bench \citep{jimenezswe} uses human-written unit tests as
evaluation metrics; Embodied Agent Interface \citep{li2024embodied} provides
fine-grained evaluation for LLM-based embodied agents; $\tau$-Bench \citep{yao2024tau}
compares database states for evaluation; concurrent work AgentRewardBench \citep{lù2025agentrewardbenchevaluatingautomaticevaluations}
builds a benchmark for reward models for web agents. Recently, there are
observatory tools including Galileo \citep{galileo_agentic}, Vertex AI Gen AI
\citep{google_agent_eval}, and Docent \citep{meng2025docent} which provide user interfaces
to visualize agent failure modes. Generating intrinsic rewards have also been
studied in the reinforcement learning community \citep{du2019liir,pathakICMl17curiosity,laskin2022cic}
to encourage exploration, sub-task completion, or skill discovery. In contrast
to these, AutoLibra is a pure data-driven task-agnostic method without predefined
failure taxonomy for generating interpretable metrics for agents.

\noindent\textbf{Learning from natural language and human feedback}
Researchers have been studying reinforcement learning with language feedback to
provide a dense reward to agents \citep{goyal2019using}. Since LLM agents are
even harder to train with sparse reward, there is substantial interest in training
LLM agents from natural language feedback. \citet{chen2024learning} propose an imitation
learning method for learning from human feedback; Text2Reward \citep{xietext2reward}
uses code generation to generate robot reward functions from open-ended human feedback;
our work \citep{chen2025fine} uses feedback to the improvement agent policy with
prompting and then align the unprompted agent policy with the prompted one;
\citet{shi2024yell} propose a new model architecture to incorporate human feedback
into policy learning. On the other hand, human non-open-ended feedback is also incorporated
in training agents, including rating feedback \citep{nguyen2017reinforcement},
preference feedback \citep{christiano2017deep}, demonstrative feedback \citep{shaikhaligning}.
Unlike these papers, AutoLibra induces metrics from feedback from all annotated instances
and generates metrics that are generalizable to different tasks and useful for both
evaluation and agent fine-tuning.

\noindent\textbf{Thematic analysis}
Thematic analysis is a powerful tool for qualitative study through coding and
iterative creation of themes. \citet{gauthier2022computational} provide
computational tools to aid this process; \citet{hong2022scholastic} and \citet{gebreegziabher2023patat}
explore human-AI collaboration in thematic analysis; LLooM \citep{lam2024concept},
an automatic method for concept induction, closly aligns with and informs our approach.
This paper completes the loop of concept induction by using the meta-evaluation
step to optimize the induced metrics, and apply it to
agent evaluation.

\section{Conclusion and Future Work}
This work introduces AutoLibra, a new paradigm for agent evaluation, one of
the first works to explore adaptable trajectory-derived evaluation heuristics,
offering substantial advantages in agent training over traditional end-to-end
evaluation. We find that this framework is generalizable to a diverse range of agent
tasks, provides new insights into agent behaviors, and identifies strong optimization
targets for agent improvement. There are a few directions for further extending
and applying this framework.
(1) \textbf{Behavior-centric evaluation} AutoLibra leads a \emph{paradigm shift}
from end-to-end agent evaluation (analogous to ``integration tests'' in software
development) to evaluation with granular metrics that measure agents' concrete behaviors
(analogous to ``unit tests''). Future work can study whether this process can be
improved through better human-AI collaboration.
(2) \textbf{Sub-trajectory feedback from humans} In AutoLibra, we label each
trajectory with one piece of feedback, and ground it into the agents' concrete
behavior which is at the sub-trajectory level. In the future, researchers can
let users directly give feedback for one or multiple steps in the trajectory, which
should lead to better feedback grounding results. Similarly, user feedback can be
collected during the interaction instead of after the agent has completed the tasks,
which is a more user-friendly way to gather high quality feedback data.
(3) \textbf{Wider exploration of agent improvement methods} In this paper, we only
explored non-parametric for agent improvement to show the utility of AutoLibra. Future work can use AutoLibra to provide dense rewards for individual
steps, and use reinforcement learning to train agents with these dense rewards.

	\section{The Bigger Picture}

	Humans learn not only from their own experience, but crucially from collective
	knowledge transmitted through social interaction \citep{tomasello1993cultural}. This
	capacity for social learning and teaching enables knowledge accumulation across
	generations, allowing us to build upon rather than reinvent skills and artifacts
	\citep{humphrey1976social}. At the core of this capability lies a distinctive
	form of intelligence that sets humans apart from species lacking cumulative culture:
	social intelligence. A fundamental question emerges: how can we leverage this social
	intelligence to enable AI agents to learn from humans and potentially teach them
	in similar ways?

	Current training paradigms for language agents predominantly focus on behavior
	cloning or reinforcement learning from self-generated experience. However, behavior
	cloning has inherent limitations for skill acquisition. First, it requires teachers
	or demonstrators to operate in the same environment as the learning agent, constraining
	the scalability of knowledge transfer. Second, agents learning from one or a few
	demonstrations may develop misunderstandings about what constitutes appropriate
	behavior. In such cases, explicit verbal feedback from teachers becomes crucial for
	establishing the boundaries of acceptable behaviors and clarifying underlying principles
	that cannot be easily inferred from demonstrations alone.

	AutoLibra offers a computational approach to address these limitations by incorporating
	the verbal feedback that humans naturally exchange in daily interactions to systematically
	improve agent behavior. Looking forward, the broader vision is to deploy AI agents
	in real-world settings where they can learn from diverse human feedback at scale.
	This approach could enable agents to acquire skills from the collective experience
	of all users, forming both general competencies and specialized skill sets at
	\emph{personal} and \emph{organizational} levels, mirroring how human communities
	accumulate and share knowledge.

	Going one step further, such agents could serve as intermediaries for accelerating
	skill and culture transmission between humans themselves---distilling expertise from
	proficient users into interpretable metrics and actionable feedback that can guide
	novices, thereby compressing traditional learning curves and democratizing access
	to specialized knowledge.


	\section*{Acknowledgments}
	This work is supported by ONR grant N000142412532, and NSF grant IIS-2247357, and DARPA grant Friction for Accountability in Conversational Transactions.
	We thank Google Cloud Platform and Modal Platform for their credits. We thank Yutong
	Zhang, Hayley Zhang, Yijia Shao, Michelle S Lam, Manling Li, Ryan Louie, Yanzhe
	Zhang, Xuhui Zhou, Maarten Sap, Sherry Tongshuang Wu, Shikhar Murty, Saujas
	Vaduguru, Chenghao Yang, Xizhi Xiao, Anant Sinha and all members of Stanford SALT
	Lab for their help and feedback throughout this project.

	\bibliography{main}
	\bibliographystyle{iclr2026_conference}

	\newpage
	\appendix
	\addtocontents{toc}{\protect
	\setcounter{tocdepth}{-1}}
	\section*{Appendix}
	
    \section*{Content of Appendix}
    \begin{itemize}
        \item[\textbf{1.}] \textbf{AutoLibra Ladder Methodology}
            \begin{itemize}
                \item \ref{appendix:illustrations} Illustration of AutoLibra Method

                \item \ref{appendix:algo1} Algorithm of AutoLibra Ladder

                \item \ref{appendix:autolibra_setup} AutoLibra Ladder Experiment
                    Configuration
            \end{itemize}

        \item[\textbf{2.}] \textbf{Baba-is-ai}
            \begin{itemize}
                \item \ref{appendix:baba_is_ai_rules} Rules and Environment Details

                \item \ref{appendix:heldout} Experiment Results

                \item \ref{appendix:babaisai} Metric Scores

                \item \ref{appendix:babaisai_metrics} Metric Examples

                \item \ref{appendix:babaisai_prompts} Prompts

                \item \ref{appendix:baba_is_ai_obs} Qualitative Observations of Agent
                    Performance
            \end{itemize}

        \item[\textbf{3.}] \textbf{MiniHack}
            \begin{itemize}
                \item \ref{appendix:minihack_rules} Rules and Environment Details

                \item \ref{appendix:heldout_mini} Experiment Results

                \item \ref{appendix:minihack} Metric Scores

                \item \ref{appendix:minihack_metrics} Metric Examples

                \item \ref{appendix:minihack_prompts} Prompts

                \item \ref{appendix:minihack_obs} Qualitative Observations of Agent
                    Performance
            \end{itemize}

        \item[\textbf{4.}] \textbf{WebVoyager}
            \begin{itemize}

                \item \ref{appendix:nnetnav_live} NNetNav-Live Induced Metrics
            \end{itemize}
    \end{itemize}

    \newpage
    \section{Illustration of AutoLibra Method}
    \label{appendix:illustrations}
    \begin{figure}[!h]
        \includegraphics[width=0.4\linewidth]{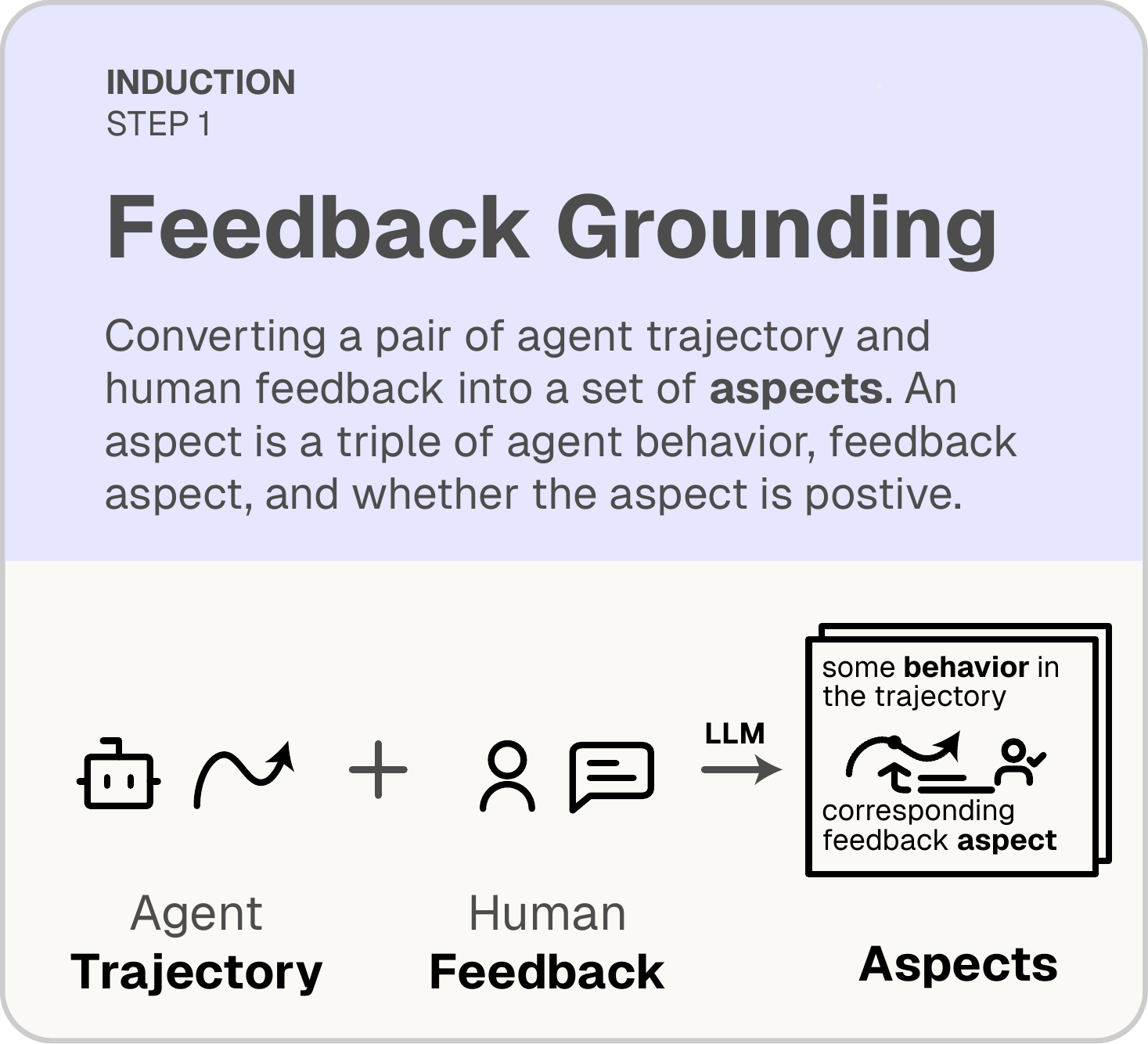}
        \caption{Feedback Grounding}
        \label{fig:feedback_grounding}
    \end{figure}

    \begin{figure}[!h]
        \includegraphics[width=0.4\linewidth]{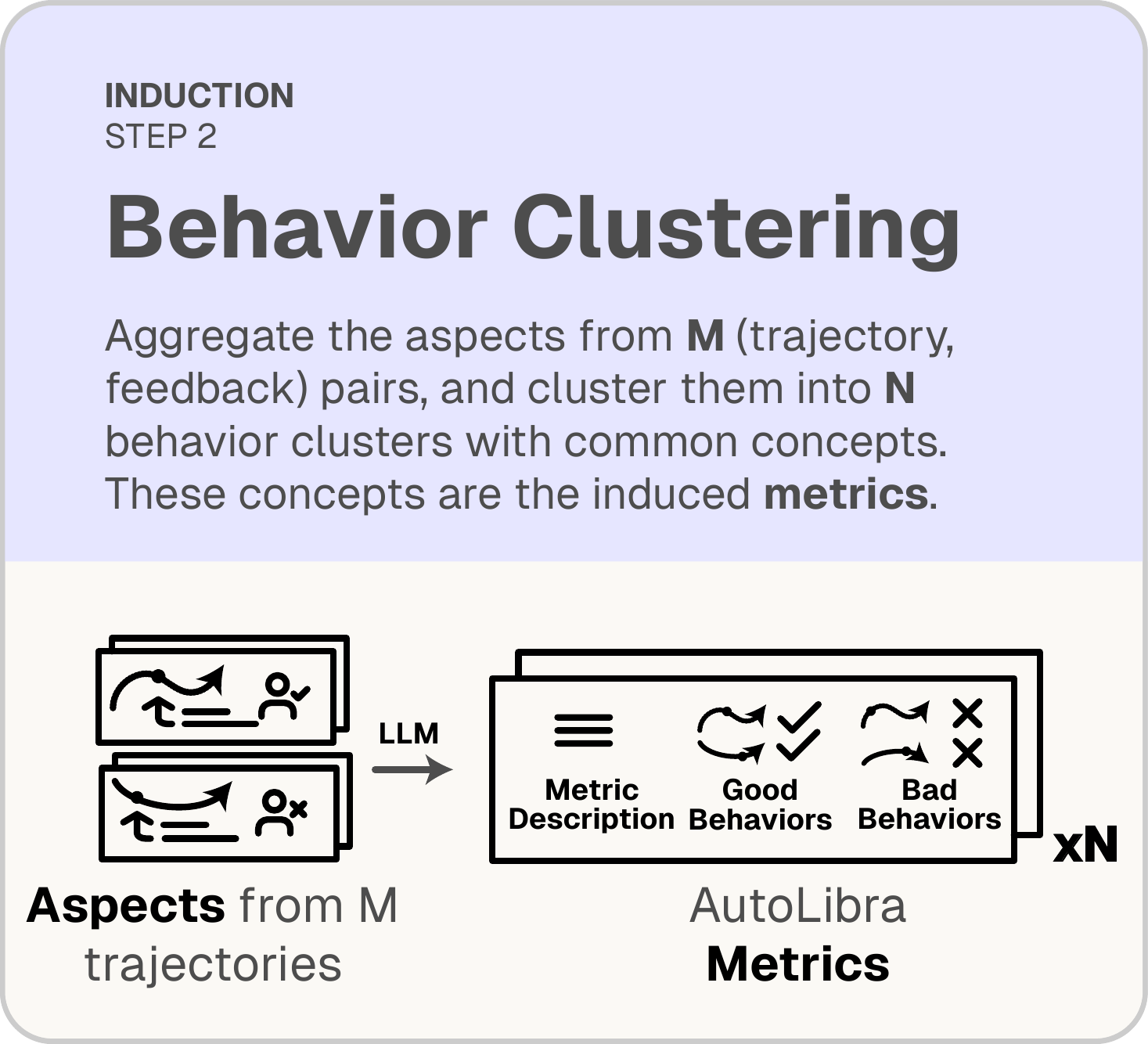}
        \caption{Behavior Clustering}
        \label{fig:behavior_clustering}
    \end{figure}

    \begin{figure}[!h]
        \includegraphics[width=0.4\linewidth]{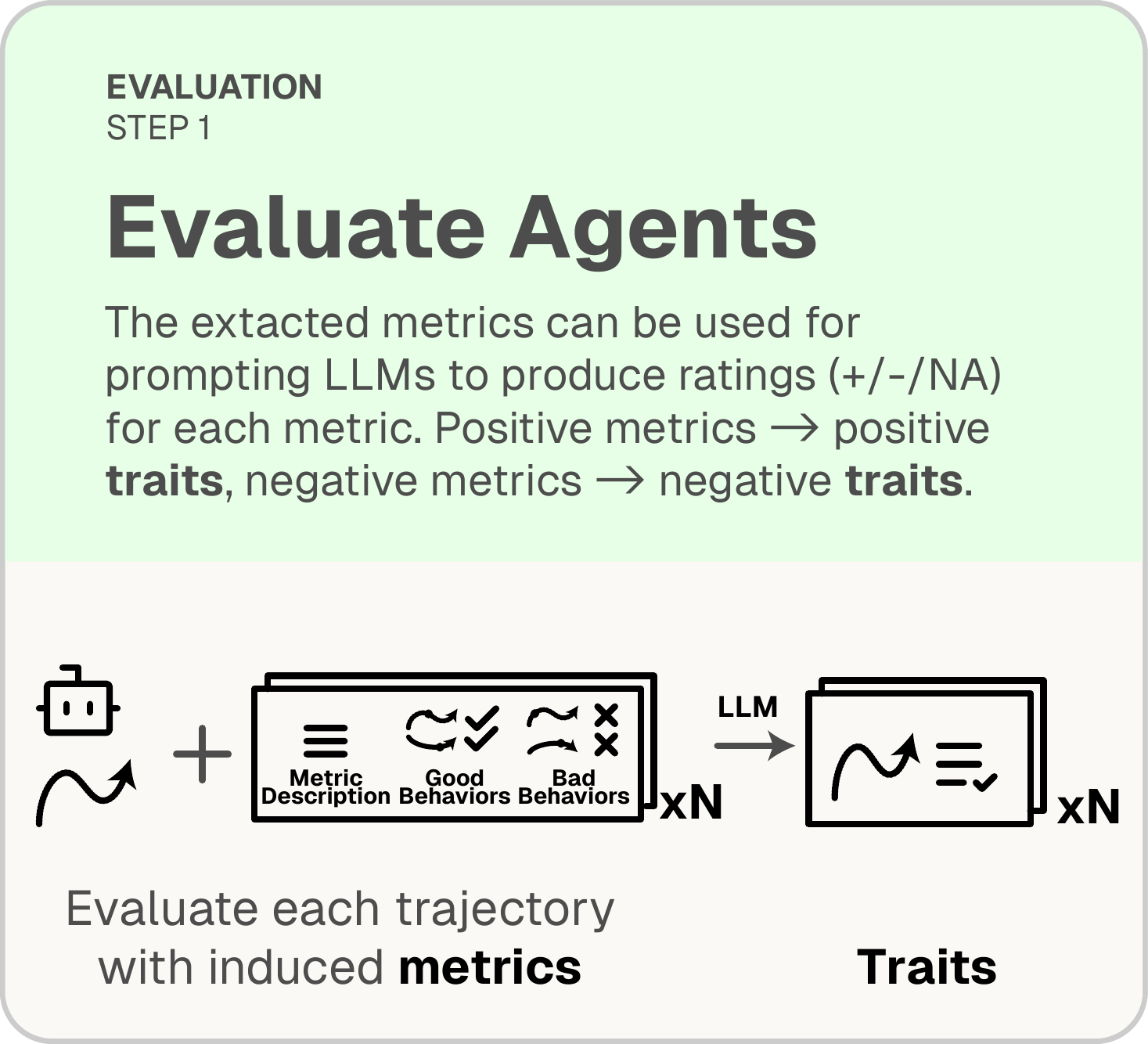}
        \caption{Evaluation with induced metrics}
        \label{fig:llm_as_a_judge}
    \end{figure}

    \begin{figure}[!h]
        \includegraphics[width=0.4\linewidth]{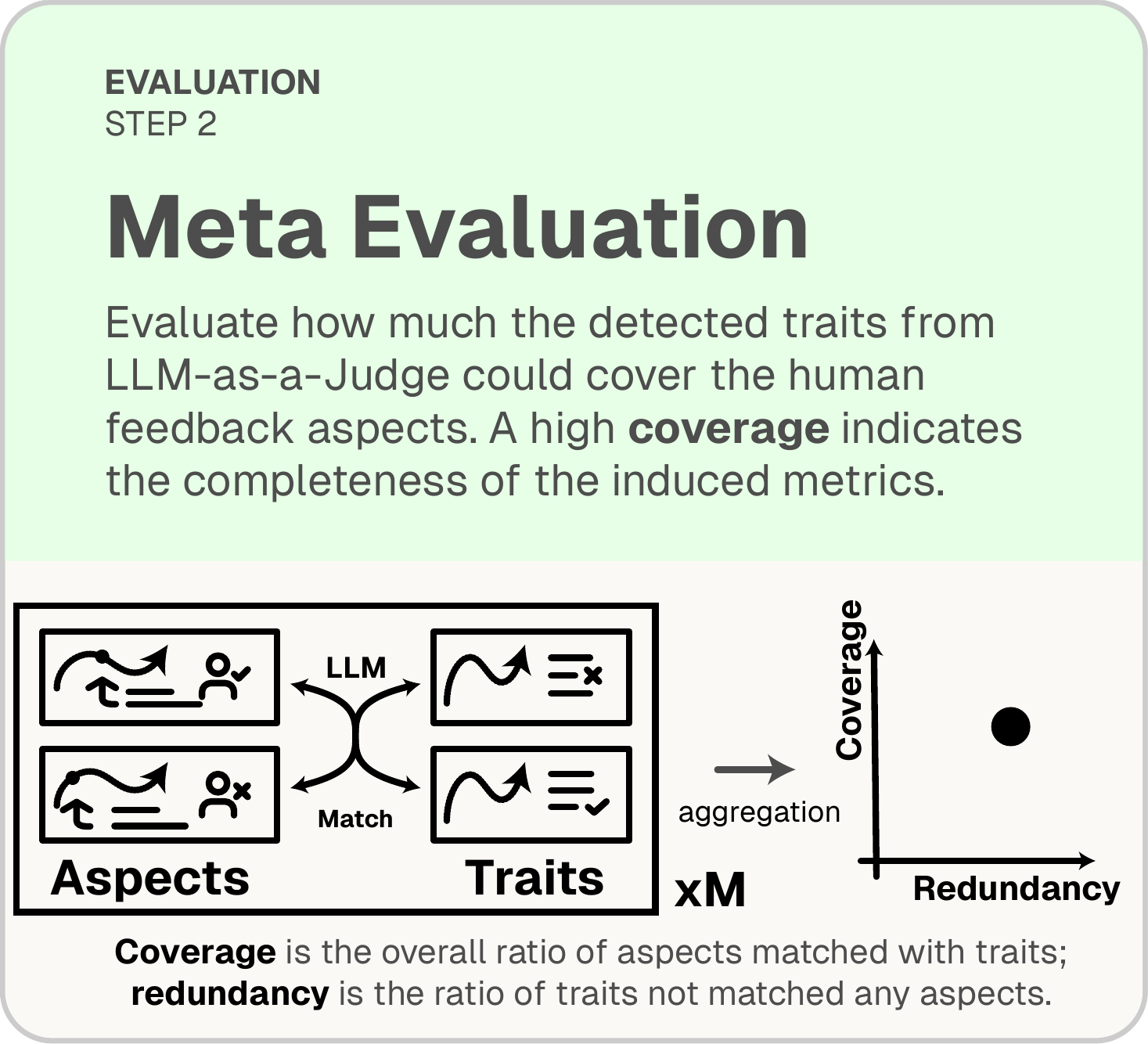}
        \caption{Meta evaluation}
        \label{fig:meta_evaluation}
    \end{figure}

    \section{Algorithm of AutoLibra Ladder Experiment}
    \label{appendix:algo1}
    \begin{algorithm}
	\caption{Pseudocode for iterative agent improvement with AutoLibra}
	\begin{algorithmic}
		[1] \For{$i$ in range($n\_iters$)} \For{$task$ in $selected\_tasks$} \State $t
		raj_{i}, eval\_score_{i}+= agent.play(task, prompt)$ \State $annotations_{i}+
		= editor.annotate(traj_{i})$ \EndFor \State $metrics_{i}= AutoLibra.extract \_
		metrics(traj_{i}, annotations_{i})$ \State $traj\_scores_{i}= AutoLibra.ll m\_
		eval(metrics_{i}, traj_{i})$ \State $curr\_scores = eval\_score_{i}$ \While {$curr\_scores <= eval\_score_{i}$}
		\State $prompt = updated\_prompt_{k}$ \For{$task$ in $selected\_tasks$}
		\State $\_, curr\_scores = agent.play(task, prompt)$ \EndFor \EndWhile
		\EndFor
	\end{algorithmic}
\end{algorithm}

    \section{AutoLibra Ladder Experiment Setup}
    \label{appendix:autolibra_setup}
    This section describes the test configuration used during AutoLibra Ladder
experiments, as described in \ref{sec:ladder}.

For each environment experiment, one unmodified agent is used as a baseline for
comparison (\textit{Iteration 0}), and three complete iterations of iterative
agent improvement with AutoLibra are performed (\textit{Iterations 1-3}), for a
total of four iterations. Six representative tasks for baba-is-ai are used in to
induce metrics within the AutoLibra pipeline, with the remaining 34 tasks held
out for evaluation. The AutoLibra Ladder pipeline is evaluated on the baba-is-ai
and MiniHack environments; any changes to the agent code, the environment score,
trajectory performance, and other metrics are recorded at the end of each iteration.
GPT-4o-241120 \cite{openai2024gpt4ocard} is used as the agent model in all
experiments; the human-in-the-loop configuration of the AutoLibra pipeline is used.

    \section{Baba-is-ai Rules and Environment Configuration}
    \label{appendix:baba_is_ai_rules}
    This section discusses the rules and implementation details of Baba is AI, the task
environment used in experiments in Section 4. Baba is AI is derived from Baba is
You, a grid-based puzzle video game, and was originally implemented as part of the
BALROG \cite{paglieri2024balrog} agent benchmark.

Baba Is You is a puzzle game where the player controls a character that can navigate
and modify the rules of the game by pushing blocks containing words such as 'you',
'is' 'key', that, when combined, define the game's rules.

The game field consists of a rectangular 4-connected grid with a number of objects,
rules blocks, and obstacles present.

\begin{figure}[ht]
	\centering
	\includegraphics[scale=0.75]{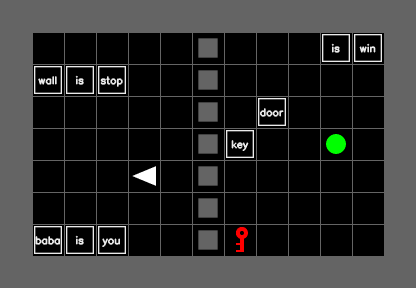}
	\caption{Example of the baba-is-ai environment. In this task (\textit{two\_room-break\_stop-make\_win-distr\_obj-irrelevant\_rule}),
	the agent has to break the "wall is stop" rule by pushing either 'wall', 'is',
	or 'stop' out of alignment with the other blocks, and must then push the "key"
	rule block next to "is win" at (9, 1) to assemble the win rule, then touch the
	key at (7, 7) to win. Pushing the "door" rule block would be a mistake, as no door
	object is present.}
	\label{fig:babaisai_env}
\end{figure}

Words (represented as rule blocks) can be combined to form sentences that define
the properties of objects, rules, and obstacles in the game, with rules becoming
active when a full sequence of rule blocks are joined in a 3-block line horizontally
or vertically.

Active rules are built in the form [<subject> <verb> <object>], where the
subject and object are the names of objects in the game, and the verb is one of the
following: "is", "has", "can", or "not".

The goal of the game is to reach a win condition, by manipulating the rules of
the game to create new win conditions, modify the properties of objects, or
change the behavior or identity of the player character.

The testing implementation we use is the baba-is-ai environment, a simplified version
of Baba is You originally implemented within the BALROG agent benchmark \cite{cloos2024babaaibreakrules,
paglieri2024balrog}. This implementation has several simplifications compared to
the original game, including a smaller grid size, fewer objects, and a limited set
of rules. The environment is designed to be easy to use and understand, while still
providing a challenging testbed for evaluating the performance of reinforcement learning
agents on agentic reasoning tasks.

A single baba-is-ai task is defined by an arbitrary rectangular grid where an exit
condition must be reached, with several possible obstacles and rules preventing
the player from reaching the exit. Intermediate goals to achieving the exit condition
can be defined by the user, and the environment will generate a task that
requires the agent to learn how to manipulate the rules of the game in order to
reach the exit. These intermediate goals are referred to as "subtasks" within our
paper, and include:

\begin{itemize}
	\item \textbf{Goto-Win}: The agent must reach a specific location on the grid.

	\item \textbf{Make-Win}: The agent must create a new win condition by manipulating
		the rules of the game.

	\item \textbf{Break-Stop}: The agent must break or bypass a wall or obstacle in
		order to reach the exit.

	\item \textbf{Change-You}: The agent must change the identity of the player character
		in order to reach the exit, by modifying the "baba is you" rule.
\end{itemize}

These subtasks can be arbitrarily combined with distractor objects or rules,
immovable game field walls, and each other to generate tasks of varying complexity
and length; baba-is-ai implements 40 total unique tasks of gradually increasing
difficulty.

The agent is provided observations in a text form, which includes the current state
of the game field, the currently active rules, and relative locations of obstacles
to the active player character in terms of shortest Manhattan distance. This is
done to make the environment compatible with purely text-based language models.

The agent is provided the space of possible actions it can take, which consist moving
one space in one of the four cardinal directions (up, down, left, right), but is
not given information about whether a given move will result in movement; it is
thus free to move into immovable walls without effect, to push rule blocks into corners
where they cannot be moved, or to make the task unsolvable by breaking the
active "baba is you" rule without taking control of a new object; in this last instance,
the environment automatically resets the task to a randomized beginning state
for that task.

The environment is limited to 100 steps per task episode to avoid tasks being
solved by random walks.

    \section{Baba-is-ai Experiment Results}
    \label{appendix:heldout}
    \renewcommand{\arraystretch}{1.5}

\begin{table}[ht] 
	\centering
	\begin{tabular}{>{\raggedright\arraybackslash}p{6cm}cccc|c}
		\toprule \textbf{Iteration}                                        & \textbf{0} & \textbf{1} & \textbf{2} & \textbf{3} & \textbf{Baseline} \\
		\midrule \textbf{Babaisai Score GPT-4o}                            & 30\%       & 40\%       & 43\%       & 55\%       & 33\%              \\
		\midrule \textbf{Babaisai Score GPT-4o (Only Held-Out)}            & 33\%       & 40\%       & 44\%       & 53\%       & 33\%              \\
		\midrule \textbf{Babaisai Score Claude 3.5 Sonnet}                 & 35\%       & 40\%       & 45\%       & 55\%       & 37\%              \\
		\midrule \textbf{Babaisai Score Claude 3.5 Sonnet (Only Held-Out)} & 38\%       & 42\%       & 47 \%      & 58\%       & 33\%              \\
		\midrule \textbf{Average Env. Steps}                               & 79         & 63         & 60         & 51         & -                 \\
		\bottomrule
	\end{tabular}
	\caption{\raggedright Baba-is-ai Scores and Average Environment Steps}
	\label{tab:heldout}
\end{table}

    \section{Baba-is-ai Metric Scores}
    \label{appendix:babaisai}
\newcommand{\cellcolorpercent}[1]{%
\ifdim #1 pt < 10 pt \cellcolor{red!90}%
\else\ifdim #1 pt < 20 pt \cellcolor{red!80}%
\else\ifdim #1 pt < 30 pt \cellcolor{red!70}%
\else\ifdim #1 pt < 40 pt \cellcolor{red!60}%
\else\ifdim #1 pt < 50 pt \cellcolor{red!50}%
\else\ifdim #1 pt < 60 pt \cellcolor{yellow!40}%
\else\ifdim #1 pt < 70 pt \cellcolor{yellow!30}%
\else\ifdim #1 pt < 80 pt \cellcolor{green!30}%
\else\ifdim #1 pt < 90 pt \cellcolor{green!50}%
\else\ifdim #1 pt < 100 pt \cellcolor{green!70}%
\else\cellcolor{green!90}%
\fi\fi\fi\fi\fi\fi\fi\fi\fi\fi }

\makeatletter
\newcommand{\thickhline}{%
\noalign {\ifnum 0=`}\fi \hrule height 3pt \futurelet \reserved@a \@xhline }
\newcolumntype{"}{@{\hskip\tabcolsep\vrule width 1pt\hskip\tabcolsep}} \makeatother

\renewcommand{\arraystretch}{1.5}
\begin{table}[ht]
	\centering
	\begin{tabular}{|>{\arraybackslash}p{5cm}|>{\arraybackslash}p{1.5cm}|c|c|c|c|}
		\hline
		\rowcolor[HTML]{C0C0C0} \textbf{}           & \textbf{Iteration}                                    & \textbf{0}                              & \textbf{1}                              & \textbf{2}                              & \textbf{3}                              \\
		\hline
		Win Condition Recognition                   & \includegraphics[scale=0.07]{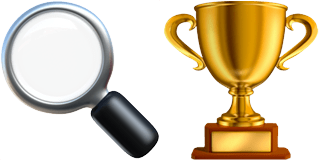} & \cellcolorpercent{35.0} \textbf{35.0\%} & \cellcolorpercent{55.0} \textbf{55.0\%} & \cellcolorpercent{87.5} \textbf{87.5\%} & \cellcolorpercent{87.5} \textbf{87.5\%} \\
		\hline
		Rule Modification                           & \includegraphics[scale=0.07]{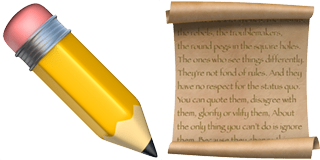} & \cellcolorpercent{0.0} \textbf{0.0\%}   & \cellcolorpercent{10.0} \textbf{10.0\%} & \cellcolorpercent{37.5} \textbf{37.5\%} & \cellcolorpercent{61.9} \textbf{61.9\%} \\
		\hline
		Direct Navigation Efficiency                & \includegraphics[scale=0.07]{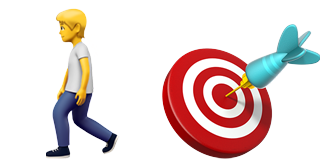} & \cellcolorpercent{5.6} \textbf{5.6\%}   & \cellcolorpercent{22.5} \textbf{22.5\%} & \cellcolorpercent{27.5} \textbf{27.5\%} & \cellcolorpercent{37.5} \textbf{37.5\%} \\
		\hline
		Context-Sensitive Decision Making           & \includegraphics[scale=0.07]{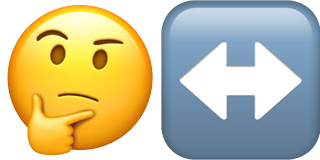} & \cellcolorpercent{2.5} \textbf{2.5\%}   & \cellcolorpercent{27.5} \textbf{27.5\%} & \cellcolorpercent{30.0} \textbf{30.0\%} & \cellcolorpercent{37.5} \textbf{37.5\%} \\
		\hline
		Win Rule Construction                       & \includegraphics[scale=0.07]{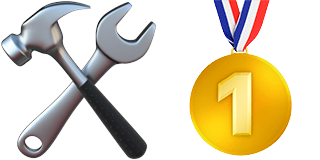} & \cellcolor[gray]{0.85} 0.0\%            & \cellcolorpercent{0.0} \textbf{0.0\%}   & \cellcolorpercent{0.0} \textbf{0.0\%}   & \cellcolorpercent{5.3} \textbf{5.3\%}   \\
		\hline
		Selective Interaction With Relevant Objects & \includegraphics[scale=0.07]{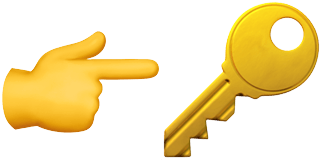} & \cellcolor[gray]{0.85} 35.0\%           & \cellcolorpercent{40.0} \textbf{40.0\%} & \cellcolorpercent{45.0} \textbf{45.0\%} & \cellcolorpercent{82.5} \textbf{82.5\%} \\
		\hline
		Rule Manipulation and Execution             & \includegraphics[scale=0.07]{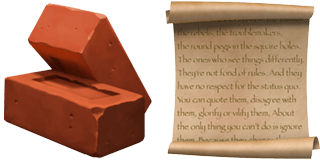} & \cellcolor[gray]{0.85} 0.0\%            & \cellcolorpercent{12.5} \textbf{12.5\%} & \cellcolorpercent{31.0} \textbf{31.0\%} & \cellcolorpercent{35.5} \textbf{35.5\%} \\
		\hline
		Subtask Coordination                        & \includegraphics[scale=0.07]{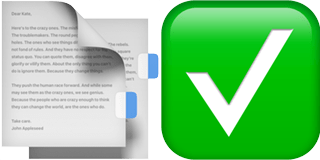} & \cellcolor[gray]{0.85} 2.5\%            & \cellcolor[gray]{0.85} 25.0\%           & \cellcolorpercent{27.5} \textbf{27.5\%} & \cellcolorpercent{35.0} \textbf{35.0\%} \\
		\hline
		Immovable Interaction                       & \includegraphics[scale=0.07]{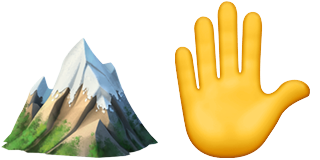} & \cellcolor[gray]{0.85} 64.9\%           & \cellcolor[gray]{0.85} 69.7\%           & \cellcolor[gray]{0.85} 69.7\%           & \cellcolorpercent{87.9} \textbf{87.9\%} \\
		\thickhline \multicolumn{2}{|c|}{Coverage}  & \textbf{65.0\%}                                       & \textbf{83.0\%}                         & \textbf{85.0\%}                         & \textbf{92.0\%}                          \\
		\hline
		\multicolumn{2}{|c|}{Redundancy}            & \textbf{58.0\%}                                       & \textbf{59.0\%}                         & \textbf{47.0\%}                         & \textbf{59.0\%}                          \\
		\hline
	\end{tabular}
	\caption{Metric Performance for baba-is-ai AutoLibra Iterations 0–3, Across Full
	(40) Environment Tasks}
	\label{tab:metric_perf}
\end{table}
    \newpage

    \section{Baba-is-ai Metric Examples}
    \label{appendix:babaisai_metrics}
    \vspace{-5mm}
\fontsize{9.5pt}{11pt}\selectfont
\begin{tcolorbox}
	[ colback=gray!5!white, colframe=gray!10, title=\textbf{\textcolor{black}{Iteration
	0: Context-Sensitive Decision Making}}]

	\textbf{Explanation:} \\ This metric assesses the agent’s capacity for context-sensitive
	decision making. It evaluates whether the agent tailors its actions according
	to the immediate game conditions—balancing between direct navigation and rule
	modification. Positive behaviors in new environments will demonstrate an ability
	to determine when obstacles require intervention and when direct movement is sufficient,
	thereby optimizing overall efficiency.

	\vspace{1em}
	\textbf{Good Behaviors:}
	\begin{itemize}
		\item Accurately gauges the game context by recognizing when obstacles are not
			an issue—such as when the win condition is already accessible—and refrains
			from unnecessary rule modifications.

		\item Selects focused, goal-oriented actions that align with observed win
			conditions, avoiding extraneous exploration.

		\item Adapts its strategy based on spatial layout and current rules,
			ensuring that its actions are timely and appropriate for the situation.
	\end{itemize}

	\vspace{0.5em}
	\textbf{Bad Behaviors:}
	\begin{itemize}
		\item Engages in excessive exploratory actions that do not contribute to reaching
			the win condition.

		\item Repeatedly takes ineffective actions (for example, persistently moving
			into walls) before finally switching strategy, indicating delayed context-sensitive
			decisions.

		\item Alters irrelevant rules or diverts attention from the active win condition
			when the situation does not demand it.

		\item Fails to adjust decision-making based on contextual cues, leading to
			uncoordinated or delayed progression toward the goal.
	\end{itemize}
\end{tcolorbox}

\fontsize{9.5pt}{11pt}\selectfont
\begin{tcolorbox}
	[ colback=gray!5!white, colframe=gray!30, title=\textbf{\textcolor{black}{Iteration
	1: Rule Modification for Obstacle Management}}]

	\textbf{Explanation:} \\ This metric measures the agent’s competence in managing
	obstacles through rule modifications. It focuses on the agent’s ability to
	detect when a game rule (such as a blocking wall or an unchangeable character assignment)
	is hindering progress and to successfully alter that rule to create a viable
	path to victory. In novel scenarios, agents displaying positive behavior will
	apply targeted rule changes that directly open the path toward the win
	condition.

	\vspace{1em}
	\textbf{Good Behaviors:}
	\begin{itemize}
		\item Proactively breaks the \texttt{'wall is stop'} rule when an obstacle
			blocks access to the win condition.

		\item Effectively modifies rules—such as replacing \texttt{'baba is you'}
			with \texttt{'key'}—to remove or bypass obstacles.
	\end{itemize}

	\vspace{0.5em}
	\textbf{Bad Behaviors:}
	\begin{itemize}
		\item Fails to modify critical rule blocks (for instance, not altering \texttt{'baba
			is you'} when required) that prevent access to the win condition.

		\item Does not interact with immovable obstacles like the \texttt{'wall is stop'}
			rule, neglecting available mechanisms to bypass them.

		\item Neglects to rearrange rule blocks to create or build necessary win conditions
			(e.g., \texttt{'door is win'}), leaving obstacles unaddressed.
	\end{itemize}
\end{tcolorbox}

\fontsize{9.5pt}{11pt}\selectfont
\begin{tcolorbox}
	[ colback=gray!5!white, colframe=gray!60, title=\textbf{\textcolor{black}{Iteration
	2: Subtask Coordination and Overall Task Planning}}]

	\textbf{Explanation:} \\ This metric assesses how well the agent coordinates multiple
	sub-tasks and plans its overall strategy. It rewards behaviors that
	demonstrate clear sequencing – from recognizing obstacles and manipulating
	rules to directly advancing toward the final objective. Failures in subtask coordination
	result in repetitive loops, ineffective transitions between actions, and an inability
	to achieve a meaningful win condition.

	\vspace{1em}
	\textbf{Good Behaviors:}
	\begin{itemize}
		\item Final movement and approach towards \texttt{'ball'} after removing the
			obstacle.

		\item Direct movement between objectives as opposed to unrelated exploration.

		\item Throughout the trajectory, the agent repeatedly chooses actions that
			reduce the distance to the door: moving left and down as needed.

		\item The trajectory demonstrates efficient movement towards the goal without
			unnecessary actions.

		\item The trajectory shows direct movement to the door without redundant backtracking
			or circular movement.

		\item Throughout the trajectory, the agent follows a direct and purposeful
			path towards its goal.
	\end{itemize}

	\vspace{0.5em}
	\textbf{Bad Behaviors:}
	\begin{itemize}
		\item The trajectory shows multiple iterations of upwards movement resulting
			in no significant progress.

		\item In the trajectory, actions such as repeated \texttt{'left'} moves where
			no significant progress towards the goal is made.

		\item In over multiple steps, the agent moves unsuccessfully against
			immovable boundaries and objects.

		\item There are periods in the trajectory where the agent exhibits loops or repetitive
			movements without advancing its position strategically.

		\item Ultimately, the agent's efforts to form victory conditions do not
			result in a meaningful or achievable goal given the map's configuration.
	\end{itemize}
\end{tcolorbox}

\fontsize{9.5pt}{11pt}\selectfont
\begin{tcolorbox}
	[ colback=gray!5!white, colframe=gray!90, title=\textbf{\textcolor{black}{Iteration
	3: Interaction with Immovable Obstacles}}]

	\textbf{Explanation:} \\ This metric measures how the agent handles immovable obstacles.
	Positive behaviors show proper recognition and effective avoidance of fixed
	objects, whereas negative behaviors involve futile or incorrect push attempts that
	betray a lack of understanding of the environment’s static features.

	\vspace{1em}
	\textbf{Good Behaviors:}
	\begin{itemize}
		\item Recognizes that immovable walls or blocks should not be pushed and instead
			plans to bypass them.

		\item Avoids colliding with immovable objects by correctly assessing their fixed
			nature.

		\item Plans actions that account for static obstacles, ensuring safe
			navigation around them.
	\end{itemize}

	\vspace{0.5em}
	\textbf{Bad Behaviors:}
	\begin{itemize}
		\item Repeatedly tries to push into an immovable wall or rule block despite the
			known constraints.

		\item Interacts with stationary obstacles in ways that disregard their immovability,
			leading to ineffective progress.

		\item Executes push commands in the wrong direction on fixed objects,
			indicating a misunderstanding of obstacle dynamics.
	\end{itemize}
\end{tcolorbox}
    \newpage

    \section{Baba-is-ai Prompts}
    \label{appendix:babaisai_prompts}
    \definecolor{lightgreen}{RGB}{220,255,220}
\definecolor{darkgreen}{RGB}{0,150,0}
\newmdenv[ backgroundcolor=lightgreen, linecolor=darkgreen, linewidth=2pt,
roundcorner=5pt, skipabove=10pt, skipbelow=10pt, leftmargin=0pt, rightmargin=0pt,
innerleftmargin=10pt, innerrightmargin=10pt, innertopmargin=10pt,
innerbottommargin=10pt, splittopskip=\topskip, splitbottomskip=10pt, frametitle={\textbf{Iteration 0 Baba-is-ai Prompt}},
frametitlebackgroundcolor=darkgreen, frametitlefont=
\color{white}
\bfseries, frametitlerule=true, frametitleaboveskip=8pt, frametitlebelowskip=8pt,
]{GreenBox}

\begin{GreenBox}
	\textbf{Baba Is You} is a puzzle game where you can manipulate the rules of
	each level. The following are the possible actions you can take in the game,
	followed by a short description of each action:

	\begin{itemize}
		\item \textbf{idle}: wait for one step,

		\item \textbf{up}: take one step up,

		\item \textbf{right}: take one step to the right,

		\item \textbf{down}: take one step down,

		\item \textbf{left}: take one step to the left.
	\end{itemize}

	\textbf{Tips:}
	\begin{itemize}
		\item Examine the level carefully, noting all objects and text blocks
			present.

		\item Identify the current rules, which are formed by text blocks in the
			format "[Subject] IS [Property]" (e.g. "BABA IS YOU").

		\item Consider how you can change or create new rules by moving text blocks around.

		\item Remember that you can only move objects or text that are not defined as
			"STOP" or similar immovable properties.

		\item Your goal is usually to reach an object defined as "WIN", but this can
			be changed.

		\item Think creatively about how changing rules can alter the properties and
			behaviors of objects in unexpected ways.

		\item If stuck, try breaking apart existing rules or forming completely new
			ones.

		\item Sometimes the solution involves making yourself a different object or changing
			what counts as the win condition.
	\end{itemize}

	\textbf{PLAY!}

	\textbf{Current Observation:}

	\textbf{Active rules:}
	\begin{itemize}
		\item ball is win

		\item baba is you
	\end{itemize}

	\textbf{Objects on the map:}
	\begin{itemize}
		\item rule \texttt{ball}: 5 steps to the left and 1 step up

		\item rule \texttt{is}: 4 steps to the left and 1 step up

		\item rule \texttt{win}: 3 steps to the left and 1 step up

		\item ball: 5 steps to the left and 2 steps down

		\item rule \texttt{baba}: 5 steps to the left and 4 steps down

		\item rule \texttt{is}: 4 steps to the left and 4 steps down

		\item rule \texttt{you}: 3 steps to the left and 4 steps down
	\end{itemize}

	First, think about the best course of action. Then, you must choose exactly
	one of the listed actions and output it strictly in the following format:

	\texttt{<|ACTION|>YOUR\_CHOSEN\_ACTION<|END|>}

	Replace \texttt{YOUR\_CHOSEN\_ACTION} with the chosen action.
\end{GreenBox}

\begin{GreenBox}
	[frametitle={\textbf{Iteration 1 Baba-is-ai Prompt}}] \label{box:baba_is_you_iter_1}
	\textbf{Baba Is You} is a puzzle game where you can manipulate the rules of
	each level. The following are the possible actions you can take in the game,
	followed by a short description of each action:

	\begin{itemize}
		\item \textbf{idle}: wait for one step,

		\item \textbf{up}: take one step up,

		\item \textbf{right}: take one step to the right,

		\item \textbf{down}: take one step down,

		\item \textbf{left}: take one step to the left.
	\end{itemize}

	\textbf{Additional Tips:}
	\begin{enumerate}
		\item The game is won by identifying a win condition and making it true by placing
			the "object is win" rule blocks next to each other in line.

		\item First, identify the win condition and where the object corresponding
			to the win condition is located.

		\item If it is blocked, identify the rules that are blocking it and try to
			remove them, or circumvent them by changing the character you control by changing
			the "baba is you" rule.

		\item If the path to the win condition is not blocked, travel directly to
			the win condition object without distractions.

		\item If a "wall is stop" rule is bounded on two sides by walls, the blocks
			cannot be moved, and you must find another way to reach the win condition.

		\item Ignore any objects not related to the win condition, as they are not
			necessary to complete the level.
	\end{enumerate}

	\textbf{Example:}

	\textit{If your observation is:}
	\begin{quote}
		Active rules: ball is win wall is stop baba is you

		Objects on the map: wall 5 steps to the right and 2 steps up \\ rule \texttt{ball}
		8 steps to the right and 2 steps up \\ rule \texttt{is} 9 steps to the right
		and 2 steps up \\ rule \texttt{win} 10 steps to the right and 2 steps up \\ rule
		\texttt{wall} 1 step up \\ rule \texttt{is} 1 step to the right and 1 step up
		\\ rule \texttt{stop} 2 steps to the right and 1 step up \\ wall 5 steps to
		the right and 1 step up \\ rule \texttt{door} 10 steps to the right and 1 step
		up \\ wall 5 steps to the right \\ ball 6 steps to the right \\ wall 5 steps
		to the right and 1 step down \\ wall 5 steps to the right and 2 steps down \\
		wall 5 steps to the right and 3 steps down \\ rule \texttt{baba} 4 steps
		down \\ rule \texttt{is} 1 step to the right and 4 steps down \\ rule \texttt{you}
		2 steps to the right and 4 steps down \\ wall 5 steps to the right and 4
		steps down \\ door 7 steps to the right and 4 steps down \\
	\end{quote}

	\textit{You should reason that:}
	\begin{itemize}
		\item The win condition is "ball is win," therefore you should reach the
			ball to win.

		\item The ball is blocked by a wall, so you should remove the "wall is stop"
			rule.

		\item The "wall is stop" rule is not bounded by walls, so you can move the
			blocks to remove the rule.

		\item The door is not necessary to reach the win condition, so you can
			ignore it.

		\item Once the "wall is stop" rule is removed, you can move directly to the
			ball to win.
	\end{itemize}

	\textbf{PLAY!}

	\textbf{Current Observation:}

	\textbf{Active rules:} baba is you

	\textbf{Objects on the map:}
	\begin{itemize}
		\item rule \texttt{is}: 2 steps to the left and 3 steps up

		\item rule \texttt{win}: 1 step to the left and 3 steps up

		\item key: 2 steps to the right and 2 steps up

		\item rule \texttt{key}: 1 step to the right and 1 step up

		\item rule \texttt{baba}: 3 steps to the left and 2 steps down

		\item rule \texttt{is}: 2 steps to the left and 2 steps down

		\item rule \texttt{you}: 1 step to the left and 2 steps down

		\item ball: 2 steps down

		\item rule \texttt{ball}: 2 steps to the right and 2 steps down
	\end{itemize}

	First, think about the best course of action. Then, you must choose exactly
	one of the listed actions and output it strictly in the following format:

	\texttt{<|ACTION|>YOUR\_CHOSEN\_ACTION<|END|>}
\end{GreenBox}
\newpage
\begin{GreenBox}
	[frametitle={\textbf{Iteration 2 Baba-is-ai Prompt}}]

	\textbf{Baba Is You} is a puzzle game where you can manipulate the rules of each
	level. The following are the possible actions you can take in the game, followed
	by a short description of each action:

	\begin{itemize}
		\item \textbf{idle}: wait for one step,

		\item \textbf{up}: take one step up,

		\item \textbf{right}: take one step to the right,

		\item \textbf{down}: take one step down,

		\item \textbf{left}: take one step to the left.
	\end{itemize}

	\textbf{You solve the puzzle by identifying the type of sub-problem, and then applying
	the following blocks of solution steps:}
	\begin{enumerate}
		\item If the win condition is not blocked and its rule is active, move to
			the win condition object.

		\item If the win condition is blocked by a wall and "wall is stop" rule is active
			and not bounded by the map boundary:
			\begin{enumerate}
				\item Move into the "wall is stop" blocks to remove the rule.

				\item Move to the win condition object.
			\end{enumerate}

		\item If the win condition is blocked by a wall and "wall is stop" rule is active
			and bounded by the map boundary:
			\begin{enumerate}
				\item Locate and move to an object rule block on your side of the wall.

				\item Push the object rule block towards the "baba is you" rule block by
					moving into it, and use this rule block to push the "baba" block out
					of the "baba is you" rule block.
			\end{enumerate}

		\item If the win condition is not active:
			\begin{enumerate}
				\item Locate the object rule block that can be pushed to the "is" rule block
					to activate the win condition.

				\item Push the object rule block to the "is" rule block, making sure
					they are adjacent.

				\item Move to the win condition object.
			\end{enumerate}

		\item If the win condition object is not present:
			\begin{enumerate}
				\item Locate an object rule block that can be pushed to the "is" rule block
					to create the win condition object.

				\item Push the object rule block to the "is" rule block, making sure
					they are adjacent.

				\item Move to the win condition object.
			\end{enumerate}
	\end{enumerate}

	\textbf{Example:}

	\textit{If your observation is:}
	\begin{quote}
		Active rules: wall is stop baba is you

		Objects on the map: wall 3 steps to the right and 3 step up \\ rule \texttt{is}
		7 steps to the right and 3 step up \\ rule \texttt{win} 8 steps to the right
		and 3 step up \\ rule \texttt{wall} 2 step to the left and 2 step up \\ rule
		\texttt{is} 1 step to the left and 2 step up \\ rule \texttt{stop} 2 step up
		\\ wall 3 steps to the right and 2 step up \\ wall 3 steps to the right and 1
		step up \\ key 4 steps to the right and 1 step up \\ wall 3 steps to the right
		\\ rule \texttt{ball} 7 steps to the right \\ wall 3 steps to the right and
		1 step down \\ wall 3 steps to the right and 2 steps down \\ rule \texttt{baba}
		2 step to the left and 3 steps down \\ rule \texttt{is} 1 step to the left
		and 3 steps down \\ rule \texttt{you} 3 steps down \\ wall 3 steps to the right
		and 3 steps down \\ ball 4 steps to the right and 3 steps down\\
	\end{quote}

	\textit{You should plan your actions as follows:}
	\begin{enumerate}
		\item The win condition is not active, so it needs to be built.

		\item The win rule can be built by pushing the "ball" block on the other side
			of the wall next to the "is" block.

		\item The "wall is stop" rule is blocking the path to the ball, so you
			should remove this rule first.

		\item The "wall is stop" rule is not bounded by walls, so you can move the
			blocks to remove the rule.
			\begin{itemize}
				\item Move 2 steps to the left and 2 steps up to reach the "wall is stop"
					rule, and push the "wall" block to remove the rule.
			\end{itemize}

		\item Once the "wall is stop" rule is removed, you can push the "ball" block
			to the "is" block to win.
			\begin{itemize}
				\item Move 7 steps to the right, 3 steps down to reach the "ball" block.

				\item Move 3 steps up to push the "ball" block to the "is" block.
			\end{itemize}

		\item Once you detect the "ball is win" rule as being active, you can move
			directly to the ball to win.
	\end{enumerate}

	\textbf{PLAY!}

	\textbf{Current Observation:}

	\textbf{Active rules:} ball is win baba is you

	\textbf{Objects on the map:}
	\begin{itemize}
		\item rule \texttt{ball}: 3 steps to the left and 3 steps up

		\item rule \texttt{is}: 2 steps to the left and 3 steps up

		\item rule \texttt{win}: 1 step to the left and 3 steps up

		\item ball: 3 steps to the left and 2 steps up

		\item key: 3 steps to the left and 1 step up

		\item rule \texttt{baba}: 3 steps to the left and 2 steps down

		\item rule \texttt{is}: 2 steps to the left and 2 steps down

		\item rule \texttt{you}: 1 step to the left and 2 steps down
	\end{itemize}

	First, think about the best course of action. Then, you must choose exactly one
	of the listed actions and output it strictly in the following format:

	\texttt{<|ACTION|>YOUR\_CHOSEN\_ACTION<|END|>}

	Replace \texttt{YOUR\_CHOSEN\_ACTION} with the chosen action.
\end{GreenBox}
\newpage

\begin{GreenBox}
	[frametitle={\textbf{Iteration 3 Baba-is-ai Prompt}}]

	\textbf{Baba Is You} is a puzzle game where you can manipulate the rules of each
	level. The following are the possible actions you can take in the game, followed
	by a short description of each action:

	\begin{itemize}
		\item \textbf{idle}: wait for one step,

		\item \textbf{up}: take one step up,

		\item \textbf{right}: take one step to the right,

		\item \textbf{down}: take one step down,

		\item \textbf{left}: take one step to the left.
	\end{itemize}

	\textbf{You solve the puzzle by identifying the type of sub-problem, and then applying
	the following blocks of solution steps:}
	\begin{enumerate}
		\item If the win condition is not blocked and its rule is active, move to
			the win condition object.

		\item If the win condition is blocked by a wall and "wall is stop" rule is active
			and not bounded by the map boundary:
			\begin{enumerate}
				\item Move into the "wall is stop" blocks to remove the rule.

				\item Move to the win condition object.
			\end{enumerate}

		\item If the win condition is blocked by a wall and "wall is stop" rule is active
			and bounded by the map boundary:
			\begin{enumerate}
				\item Locate and move to an object rule block on your side of the wall.

				\item Push the object rule block towards the "baba is you" rule block by
					moving into it, and use this rule block to push the "baba" block out
					of the "baba is you" rule block.
			\end{enumerate}

		\item If the win condition is not active:
			\begin{enumerate}
				\item Locate the object rule block that can be pushed to the "is" rule block
					to activate the win condition.

				\item Push the object rule block to the "is" rule block, making sure
					they are adjacent.

				\item Move to the win condition object.
			\end{enumerate}

		\item If the win condition object is not present:
			\begin{enumerate}
				\item Locate an object rule block that can be pushed to the "is" rule block
					to create the win condition object.

				\item Push the object rule block to the "is" rule block, making sure
					they are adjacent.

				\item Move to the win condition object.
			\end{enumerate}
	\end{enumerate}

	\textbf{Example:}

	\textit{If your observation is:}
	\begin{quote}
		\textbf{Active rules:} \\ wall is stop \\ baba is you \\

		\textbf{Objects on the map:} \\ rule \texttt{wall} at (1, 1) \\ rule \texttt{is}
		at (2, 1) \\ rule \texttt{stop} at (3, 1) \\ rule \texttt{is} at (7, 3) \\
		rule \texttt{win} at (8, 3) \\ rule \texttt{ball} at (7, 0) \\ rule \texttt{baba}
		at (5, 6) \\ rule \texttt{is} at (6, 6) \\ rule \texttt{you} at (7, 6) \\
		wall at (4, 0), (4, 1), (4, 2), (4, 3), (4, 4), (4, 5), (4, 6) \\ key at (5,
		1) \\ ball at (5, 6) \\

		\textbf{Current position:} (4, 3) in a grid of shape (1, 0) to (8, 6)
	\end{quote}

	\textit{You should plan your actions as follows:}
	\begin{enumerate}
		\item The win condition is not active, so it needs to be built.

		\item The win rule can be built by pushing the \texttt{ball} block on the
			other side of the wall next to the \texttt{is} block.

		\item The "wall is stop" rule is blocking the path to the ball, so you
			should remove this rule first.

		\item The "wall is stop" rule is not bounded by walls, so you can move the
			blocks to remove the rule.
			\begin{itemize}
				\item Move 2 steps to the left and 2 steps up to reach the "wall is stop"
					rule, and push the \texttt{wall} block to remove the rule.
			\end{itemize}

		\item Once the rule is removed, you can push the \texttt{ball} block to the \texttt{is}
			block.
			\begin{itemize}
				\item Move 7 steps to the right and 3 steps down to reach the \texttt{ball}
					block.

				\item Move 3 steps up to push the \texttt{ball} block to the \texttt{is}
					block.
			\end{itemize}

		\item Once you detect the \texttt{ball is win} rule is active, move directly
			to the ball to win.
	\end{enumerate}

	\textbf{PLAY!}

	\textbf{Current Observation:}

	\textbf{Active rules:} \\ baba is you \\

	\textbf{Objects on the map:}
	\begin{itemize}
		\item rule \texttt{is}: at (2, 1)

		\item rule \texttt{win}: at (3, 1)

		\item key: at (6, 2)

		\item rule \texttt{key}: at (3, 4)

		\item door: at (6, 5)

		\item rule \texttt{baba}: at (1, 6)

		\item rule \texttt{is}: at (2, 6)

		\item rule \texttt{you}: at (3, 6)

		\item rule \texttt{door}: at (6, 6)
	\end{itemize}

	\textbf{Current position:} (4, 3) in a grid of shape (1, 1) to (6, 6)

	\textbf{Rule block pushing physics explanation:}

	\begin{itemize}
		\item Avoid moving into any non-wall rule block unless you are ready to interact
			with it.

		\item Do not push "wall is stop" blocks if the "wall is stop" rule is not currently
			active.
	\end{itemize}

	\textbf{Steps to avoid unwanted interaction:}
	\begin{enumerate}
		\item Read all rule block positions.

		\item Read your current position and direction.

		\item If a rule block is in your path, change direction by 90° (e.g., if going
			up, go left or right).

		\item If a rule is on the wall of the grid, it can only be pushed parallel
			to that wall.
	\end{enumerate}

	\textbf{Example:}

	If you are at (3, 3) and a rule block is at (3, 4), you cannot push it up by first
	moving down. Instead, move to (2, 3) or (4, 3), then move down to (3, 6) to push
	it upward.

	\textbf{Strategic Planning:}

	First, think about your current goal based on the game state. If an active rule
	changes, your goal likely changes too.

	\textbf{Most Important Rule:} If "wall is stop" is active, find a way to
	remove it or gain control of a different object before doing anything else.

	\textbf{Pushing Strategy:}
	\begin{itemize}
		\item Always push blocks upward first (maximum 10 spaces), then push them
			left or right.

		\item Rule \texttt{blocks} are written in backticks. Objects are not. Only
			\texttt{blocks} can be pushed.

		\item Never interact directly with the \texttt{is} or \texttt{win} blocks.
	\end{itemize}

	\textbf{Rule Construction Example:}

	If \texttt{win} is at (11, 1) and \texttt{is} is at (10, 1), place the object rule
	at (9, 1) to form a valid rule.

	\textbf{Coordinate System:}
	\begin{itemize}
		\item Top-left is (1, 1), bottom-right is (x, y).

		\item Moving right increases x, moving down increases y.
	\end{itemize}

	\textbf{Goal Planning Example:}

	\textbf{Active rules:}
	\begin{itemize}
		\item wall is stop — Must be broken under all circumstances!
	\end{itemize}

	\textbf{Steps to achieve the goal:}
	\begin{enumerate}
		\item Break "wall is stop" by moving into its blocks from below.

		\item Build "ball is win" rule.

		\item \texttt{ball} block is at (7, 4), \texttt{is} is at (8, 1). So:
			\begin{itemize}
				\item \texttt{ball} needs to move 3 steps up, 1 step right.

				\item Push from the left (1 step) and from below (3 steps).
			\end{itemize}

		\item If you're at (6, 4), move 1 step down and 1 step right to reach (7, 5),
			then push from below.

		\item Move 3 steps up.
	\end{enumerate}

	Then, you must choose exactly one of the listed actions and output it strictly
	in the following format:

	\texttt{<|ACTION|>YOUR\_CHOSEN\_ACTION<|END|>}
\end{GreenBox}
    \newpage

    \section{Qualitative Observations of baba-is-ai Agent Performance}
    \label{appendix:baba_is_ai_obs}
    The metrics induced by AutoLibra capture the behavior of the agent effectively, with
coverage increasing from 65\% at Iteration 0 to 92\% at \emph{Iteration 3} and a
low mean redundancy of 56\%. Notably, earlier metrics were found to describe higher-level
behaviors than later metrics, reflective of the more complex behaviors that the
agent demonstrated in later iterations, as well as the compositionality of induced
metrics. Code changes were selected to specifically target given metrics, and the
agent's performance on the targeted metric improved significantly in the iteration
following the code change. This demonstrates the utility of AutoLibra for fine-grained
agent improvement, as well as the human interpretability of the induced metrics.

The section below discusses iteration-by-iteration induced metrics, code changes
directed by the metrics, and the results of these code changes on agent environment
and trajectory performance.

\textbf{Iteration 0} The agent behavior is stochastic and inconsistent, with no clear
strategy or goal. Any progress in the environment is the consequence of a random
walk. \texttt{Win Condition Recognition}
\includegraphics[scale=0.07]{figs/emojis/emoji_1.png}
was identified as a prerequisite to all other metrics induced in this iteration,
as progress in the environment is impossible without recognizing the win condition,
and was therefore targeted for improvement. Improvements took the form of few-shot
prompting, by supplying an example of observations and the corresponding win
condition and subtasks, and guidance on subtask-level planning, by mapping
specific environmental observations to actions that would progress the agent
towards the win condition. \textbf{Iteration 1} An increase in
\includegraphics[scale=0.07]{figs/emojis/emoji_1.png}
performance of 22\% was observed between \emph{Iteration 0} and \emph{Iteration
1}, indicating that the code changes were effective in improving the agent's
ability to recognize the win condition. A slight but statistically insignificant
increase was observed for the other metrics induced in \emph{Iteration 0}, but the
agent's overall baba-is-ai environment score increased by 10\%, indicating that
\includegraphics[scale=0.07]{figs/emojis/emoji_1.png}
was correctly identified by AutoLibra as a key bottleneck to the agent's performance
in the environment.

The agent now targets specific blocks and objects instead of moving randomly,
but gets stuck in loops and on immovable objects, and is unable to consistently complete
multi-step tasks.

Based on the metrics induced in \emph{Iteration 1}, several changes to the agent
code were implemented:
\begin{itemize}
	\item Augmentation of existing subtask-level planning guidance with low-level position
		instructions, targeting improvement of \texttt{Rule Modification for
		Obstacle Management}
		\includegraphics[scale=0.07]{figs/emojis/emoji_2.png}
		and \texttt{Direct Navigation Efficiency}
		\includegraphics[scale=0.07]{figs/emojis/emoji_3.png}

	\item Meta-prompting \cite{Suzgun_Kalai_2024} by providing subtask identification
		heuristic, targeting improvement of \texttt{Selective Interaction with
		Relevant Objects}
		\includegraphics[scale=0.07]{figs/emojis/emoji_6.png}
		and \texttt{Context-Sensitive Decision Making}
		\includegraphics[scale=0.07]{figs/emojis/emoji_4.png}

	\item Augmentation of single-shot example with movement instructions,
		targeting improvement of \texttt{Rule Manipulation and \\Execution}
		\includegraphics[scale=0.07]{figs/emojis/emoji_7.png}
\end{itemize}

\textbf{Iteration 2} Substantial improvements in
\includegraphics[scale=0.07]{figs/emojis/emoji_1.png}
were observed due to more detailed single-shot examples and reasoning templates,
and the corresponding increase in other metrics supports idea that
\includegraphics[scale=0.07]{figs/emojis/emoji_1.png}
is a key bottleneck to the agent's performance. We further observed a
significant increase in
\includegraphics[scale=0.07]{figs/emojis/emoji_2.png}
and
\includegraphics[scale=0.07]{figs/emojis/emoji_7.png}
, indicating that the changes targeting these metrics were effective in
improving the agent's reasoning performance and consequent avoidance of
irrelevant objects. \texttt{Subtask Identification}
\includegraphics[scale=0.07]{figs/emojis/emoji_5.png}
saw no significant improvement, as the agent still misunderstands map directions
and confuses rule blocks with objects.

Qualitatively, the agent now recognizes when wall rules need to be broken and
breaks them, but still cannot assemble rules to win, as it gets stuck on
immovable blocks and walls and does not understand where rule blocks need to be
placed to form a win condition.

Based on the metrics induced in \emph{Iteration 2}, several changes to the agent
code were implemented:
\begin{itemize}
	\item Formatting observations as absolute position (as opposed to relative
		steps from the agent), and listing of immovable/movable blocks to improve
		\texttt{Direct Navigation Efficiency}
		\includegraphics[scale=0.07]{figs/emojis/emoji_3.png}

	\item Chain-of-thought reasoning to assemble subtask completion agenda based on
		iteration-to-iteration observations, targeting improvement of \texttt{Subtask
		Coordination and Overall Task Planning}
		\includegraphics[scale=0.07]{figs/emojis/emoji_8.png}

	\item Augmentation of single-shot example to few-shot example with reasoning templates,
		targeting improvement of \texttt{Rule Manipulation and Execution}
		\includegraphics[scale=0.07]{figs/emojis/emoji_7.png}

	\item Explicit instructions on navigation to avoid block collisions and assembly
		of rules, including movement templates, targeting improvement of \texttt{Rule
		Manipulation and Execution}
		\includegraphics[scale=0.07]{figs/emojis/emoji_7.png}
		and \texttt{Win Rule Construction}
		\includegraphics[scale=0.07]{figs/emojis/emoji_5.png}
\end{itemize}

\textbf{Iteration 3} We observe a large increase in
\includegraphics[scale=0.07]{figs/emojis/emoji_4.png}
and
\includegraphics[scale=0.07]{figs/emojis/emoji_6.png}
, indicating that the changes targeting these metrics were effective in improving
the agent's reasoning performance and consequent avoidance of irrelevant objects.
\includegraphics[scale=0.07]{figs/emojis/emoji_3.png}
saw further improvement, indicating that the use of absolute position and immovable/movable
block listing was effective in improving the agent's navigation efficiency.
\includegraphics[scale=0.07]{figs/emojis/emoji_7.png}
saw a slight increase, indicating that the changes targeting this metric were effective
in improving the agent's ability to manipulate rules to achieve the win condition.
\includegraphics[scale=0.07]{figs/emojis/emoji_5.png}
saw its first improvement, indicating that more complex metrics are effectively targeted
when the base performance and simpler metrics are improved.

Qualitatively, nearly every agent always understands when wall rules can be and
cannot be broken, as well as when it's not necessary to break the wall rule,
covers nearly all tasks that involve going to win. The agent understands how to assemble
win rules but still struggles with changing block direction and understanding that
blocks get stuck against walls.
    \newpage

    \section{MiniHack Rules and Environment Configuration}
    \label{appendix:minihack_rules}
    \begin{figure}[ht]
	\centering
	\includegraphics[width=\textwidth]{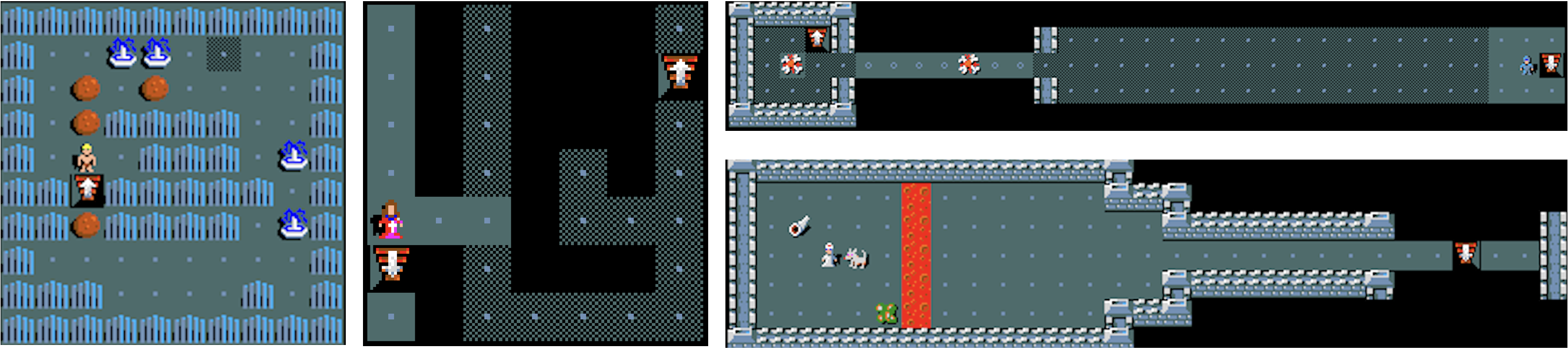}
	\caption{Overview of representative tasks used in MiniHack agent optimization
	process with AutoLibra. From left to right, up to down: (1) \emph{Boxoban}, (2)
	\emph{MazeWalk}, (3) \emph{Corridor Fight}, and (4) \emph{Quest}.}
	\label{fig:minihack_maps}
\end{figure}

This section discusses the rules and implementation details of MiniHack, another
task environment used in experiments in Section 4. Similarly to Baba is AI, MiniHack
is derived from a grid-based puzzle video game (NetHack), and was originally
implemented as part of the BALROG \cite{paglieri2024balrog} agent benchmark.

MiniHack is a grid navigation game expressed in text similar to baba-is-ai, consisting
of a procedurally generated environment that requires the agent to navigate a space
consisting of various agent roles, creatures, items, and tasks to reach a goal
\cite{samvelyan2021minihackplanetsandboxopenended}. Given its plasticity and abundant
elements, MiniHack is more complex, challenging, and diversified than baba-is-ai;
this is reflected in a lower success rate for agents on MiniHack versus Baba is AI,
with a baseline agent task completion rate of 10\% on MiniHack vs 33\% on baba-is-ai
\cite{paglieri2024balrog}.

Similar to our experiments with baba-is-ai, the Ladder improvement process for MiniHack
also follows the algorithm detailed in Appendix \ref{appendix:algo1}. Two full
iterations of agent improvement with AutoLibra were performed on MiniHack. Four representative
tasks for MiniHack are used in iterative metric improvement, with the remainder
held out for evaluation. The agent is evaluated on the MiniHack environment and any
changes to the agent code at the beginning of each iteration, and the
environment score, trajectory performance, and other metrics are recorded at the
end of each iteration. GPT-4o-241120 is used as the agent model.\\

The four selected representative tasks each contain a unique subset of subtasks
evaluating an agent's capability. Figure \ref{fig:minihack_maps} shows an example
for each task, and from left to right, up to down, these maps respectively
represent Boxoban, MazeWalk, Corridor Fight, and Quest.

\textbf{Boxoban} is a box-pushing puzzle game inspired by Sokoban, rendered within
the MiniHack environment. To succeed in Boxoban, the agent needs to push the four
boulders (orange balls) onto the four fountains (blue icons), and partial credit
will be awarded for pushing some of the boulders onto the fountains. Boxoban tests
the agent's capability in strategic planning and rule-following.
\newline
\\ \textbf{MazeWalk} is a game that requires the agent to explore unknown dark spaces
to find the target exit staircase (the icon with a downward arrow). Two
challenges for MazeWalk are fog of war -- the agent initially lacks information about
areas of the map it hasn't visited, and must explore to discover the map and
maze layout, and darkness -- even if the agent has visited a block, the block will
become 'dark' as the agent walks away and it passes out of view, retaining
information about its layout but not any enemies or items present. Thus,
MazeWalk tests the agent's capability in map memory and strategic searching.
\newline
\\ \textbf{Corridor Fight} is a game that requires the agent to explore an
unknown dark corridor map to find the target exit staircase while engaging or
avoiding giant rat enemies. Corridor Fight tests the agent's capability in memory,
space awareness, hazard awareness, and strategic combat.
\newline
\\ \textbf{Quest} requires the agent to use a tool to help itself cross an
otherwise impassable wall of lava, survive randomly generated monsters, and
search for the target exit staircase. As the most subtask-rich and randomized game,
it tests the agent's abilities to recognize and utilize tools, understand its role
and special power, and strategically survive from monsters.

In all tasks, the agent is provided observations in a text form, which includes
the current state of the game field, the currently active rules, and relative
locations of obstacles to the active player character in terms of shortest
Manhattan distance. This is done to make the environment compatible with purely text-based
language models.

The environment is limited to 100 steps per task episode to avoid tasks being solved
by random walks.

    \section{MiniHack Experiment Results}
    \label{appendix:heldout_mini}
    \begin{table}[h]
	\centering
	\renewcommand{\arraystretch}{1.5}
	\begin{tabular}{p{4cm}ccc|c}
		\toprule \textbf{Turn}                  & \textbf{0} & \textbf{1} & \textbf{2} & \textbf{Baseline} \\
		\midrule \textbf{MiniHack Score GPT-4o} & 0\%        & 12.5\%     & 25\%       & 10\%              \\
		\midrule \textbf{Average Env. Steps}    & 85         & 91         & 88         & -                 \\
		\bottomrule
	\end{tabular}
	\caption{MiniHack Score and Average Environment Steps}
	\label{tab:heldout_mini}
\end{table}

    \section{MiniHack Metric Performance}
    \label{appendix:minihack}
    \renewcommand{\arraystretch}{1.5}
\begin{table}[ht]
	\centering
	\begin{tabular}{|>{\arraybackslash}p{6cm}|>{\arraybackslash}p{1.5cm}|c|c|c|}
		\hline
		\rowcolor[HTML]{C0C0C0} \textbf{}                & \textbf{Iteration}                                   & \textbf{0}                                & \textbf{1}                                & \textbf{2}                                \\
		\hline
		Target Navigation Effectiveness                  & \includegraphics[scale=0.09]{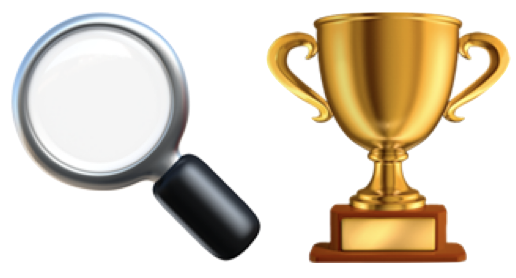} & \cellcolorpercent{16.67} \textbf{16.67\%} & \cellcolorpercent{8.33} \textbf{8.33\%}   & \cellcolorpercent{41.67} \textbf{41.67\%} \\
		\hline
		Efficient Exploration and Map Memory Utilization & \includegraphics[scale=0.07]{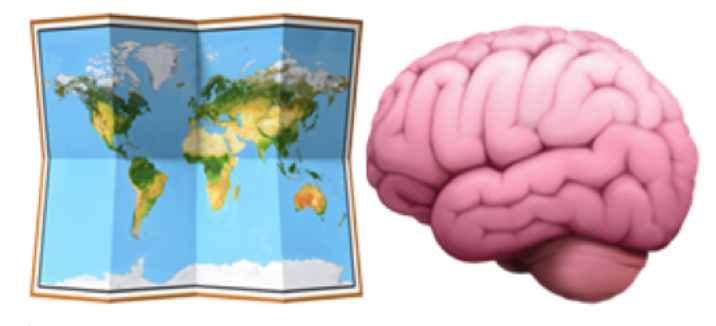} & \cellcolorpercent{16.67} \textbf{16.67\%} & \cellcolorpercent{0.00} \textbf{0.00\%}   & \cellcolorpercent{25.00} \textbf{25.00\%} \\
		\hline
		Hazard Awareness and Equipment Utilization       & \includegraphics[scale=0.09]{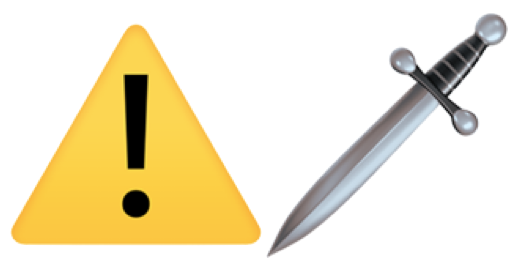} & \cellcolorpercent{0.00} \textbf{0.00\%}   & \cellcolorpercent{0.00} \textbf{0.00\%}   & \cellcolorpercent{0.00} \textbf{0.00\%}   \\
		\hline
		Boulder Manipulation Strategy                    & \includegraphics[scale=0.09]{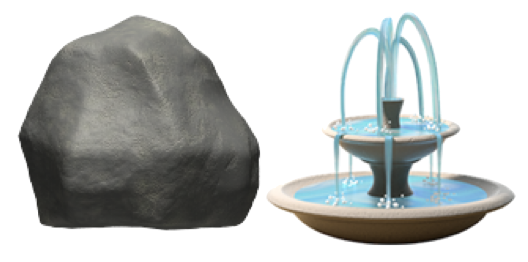} & \cellcolorpercent{0.00} \textbf{0.00\%}   & \cellcolorpercent{0.00} \textbf{0.00\%}   & \cellcolorpercent{0.00} \textbf{0.00\%}   \\
		\hline
		Combat Engagement and Survival                   & \includegraphics[scale=0.09]{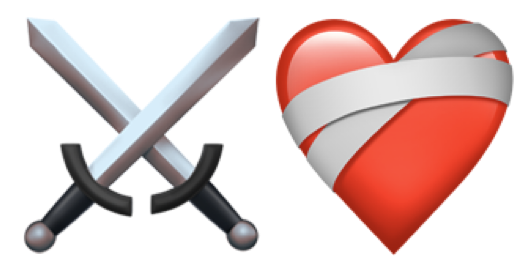} & \cellcolorpercent{8.33} \textbf{8.33\%}   & \cellcolorpercent{8.33} \textbf{8.33\%}   & \cellcolorpercent{25.00} \textbf{25.00\%} \\
		\hline
		Role-Specific Ability Utilization                & \includegraphics[scale=0.07]{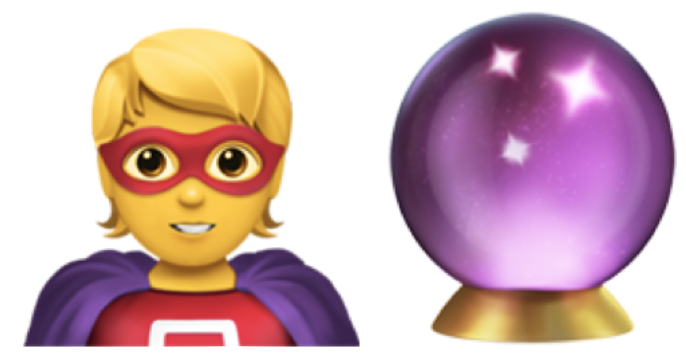} & \cellcolorpercent{0.00} \textbf{0.00\%}   & \cellcolorpercent{0.00} \textbf{0.00\%}   & \cellcolorpercent{0.00} \textbf{0.00\%}   \\
		\hline
		Spatial Awareness and Interpretation             & \includegraphics[scale=0.08]{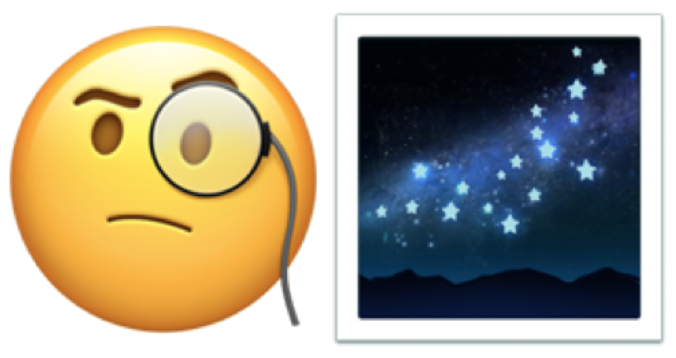} & {-}                                       & \cellcolorpercent{16.67} \textbf{16.67\%} & \cellcolorpercent{58.33} \textbf{58.33\%} \\
		\hline
		Object Pickup Efficiency                         & \includegraphics[scale=0.08]{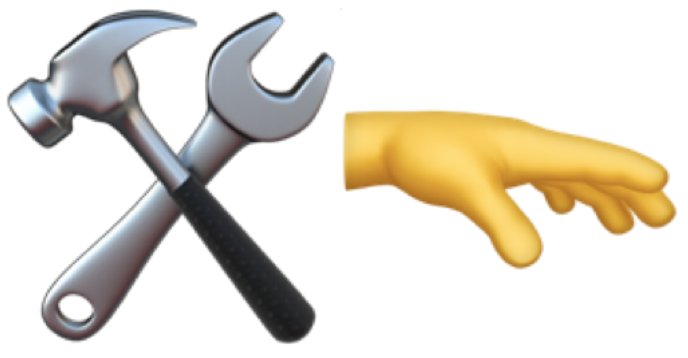} & {-}                                       & \cellcolorpercent{0.00} \textbf{0.00\%}   & \cellcolorpercent{0.00} \textbf{0.00\%}   \\
		\hline
		Giant Rats Encounter Handling                    & \includegraphics[scale=0.08]{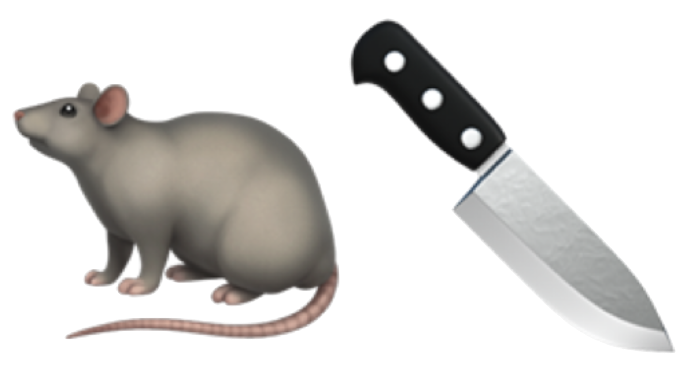} & {-}                                       & {-}                                       & \cellcolorpercent{25.00} \textbf{25.00\%} \\
		\thickhline                                       
		\multicolumn{2}{|c|}{Coverage}                   & 82.89\%                                              & 81.82\%                                   & 87.84\%                                    \\
		\hline
		\multicolumn{2}{|c|}{Redundancy}                 & 61.11\%                                              & 65.63\%                                   & 71.30\%                                    \\
		\hline
	\end{tabular}
	\caption{Metric Performance for MiniHack AutoLibra Iterations 0–2}
	\label{tab:metric_mini_perf}
\end{table}
    \newpage

    \section{MiniHack Metric Examples}
    \label{appendix:minihack_metrics}
    \fontsize{9.5pt}{11pt}\selectfont
\begin{tcolorbox}
	[ colback=gray!10!white, colframe=gray!50!black, title=\textbf{Iteration 0: Target
	Navigation Effectiveness}]

	\textbf{Explanation:} \\ This metric evaluates the agent’s ability to accurately
	identify and navigate toward key goal locations (e.g., the downward stairs).
	Positive behaviors show clear planning and directional focus, while negative examples
	reveal distractions or failures in goal-oriented movement.

	\vspace{1em}
	\textbf{Good Behaviors:}
	\begin{itemize}
		\item Behavior: The agent decides to move south and reaches the downward
			stairs.

		\item Behavior: The agent’s reasoning shows it plans its action to reach the
			stairs.

		\item Behavior: From its reasoning, the agent appears to recognize the goal of
			locating the stairs downward.

		\item Behavior: From the trajectory and reasoning, the agent consistently aimed
			toward the general direction of its goal—the stairs down.
	\end{itemize}

	\vspace{0.5em}
	\textbf{Bad Behaviors:}
	\begin{itemize}
		\item Behavior: The repeated attempts to kick the goblin instead of
			progressing towards the stairs.

		\item Behavior: The agent repeatedly interacts with adjacent stairs up but
			does not visibly attempt to identify or locate downward stairs.

		\item Behavior: The agent explored the map but did not identify the downward
			stair in its trajectory.

		\item Behavior: Throughout the trajectory, the agent did not manage to locate
			the downward stairs, despite exploring various places in the environment.

		\item Behavior: The agent spends multiple actions traversing the map without
			making meaningful progress toward the downward stairs.

		\item Behavior: Performed in the sequence of actions where the agent
			explores but eventually doesn't find the stairway downward, ending the session
			unsuccessfully.

		\item Behavior: During its navigation, the agent didn't succeed in reaching the
			stairs down.
	\end{itemize}
\end{tcolorbox}
    \newpage

    \section{MiniHack Prompts}
    \label{appendix:minihack_prompts}
    \definecolor{lightgreen}{RGB}{220,255,220}
\definecolor{darkgreen}{RGB}{0,150,0}
\newmdenv[ backgroundcolor=lightgreen, linecolor=darkgreen, linewidth=2pt,
roundcorner=5pt, skipabove=10pt, skipbelow=10pt, leftmargin=0pt, rightmargin=0pt,
innerleftmargin=10pt, innerrightmargin=10pt, innertopmargin=10pt,
innerbottommargin=10pt, splittopskip=\topskip, splitbottomskip=10pt, frametitle={\textbf{Iteration 0 MiniHack Prompt}},
frametitlebackgroundcolor=darkgreen, frametitlefont={\color{white}\bfseries},
frametitlerule=true, frametitleaboveskip=8pt, frametitlebelowskip=8pt, ]{MyGreenBox}
\BeforeBeginEnvironment{MyGreenBox}{\VerbatimEnvironment\small}

\begin{MyGreenBox}
	[frametitle={\textbf{Iteration 0 MiniHack Prompt}}] You are an agent playing MiniHack.
	The following are the possible actions you can take in the game, followed by a
	short description of each action:

	\begin{itemize}
		\item \textbf{north}: move north,

		\item \textbf{east}: move east,

		\item \textbf{south}: move south,

		\item \textbf{west}: move west,

		\item \textbf{northeast}: move northeast,

		\item \textbf{southeast}: move southeast,

		\item \textbf{southwest}: move southwest,

		\item \textbf{northwest}: move northwest,

		\item \textbf{far north}: move far north,

		\item \textbf{far east}: move far east,

		\item \textbf{far south}: move far south,

		\item \textbf{far west}: move far west,

		\item \textbf{far northeast}: move far northeast,

		\item \textbf{far southeast}: move far southeast,

		\item \textbf{far southwest}: move far southwest,

		\item \textbf{far northwest}: move far northwest,

		\item \textbf{up}: go up the stairs,

		\item \textbf{down}: go down the stairs,

		\item \textbf{wait}: rest one move while doing nothing,

		\item \textbf{more}: display more of the message,

		\item \textbf{apply}: apply (use) a tool,

		\item \textbf{close}: close an adjacent door,

		\item \textbf{open}: open an adjacent door,

		\item \textbf{eat}: eat something,

		\item \textbf{force}: force a lock,

		\item \textbf{kick}: kick an enemy or a locked door or chest,

		\item \textbf{loot}: loot a box on the floor,

		\item \textbf{pickup}: pick up things at the current location if there are any,

		\item \textbf{pray}: pray to the gods for help,

		\item \textbf{puton}: put on an accessory,

		\item \textbf{quaff}: quaff (drink) something,

		\item \textbf{search}: search for hidden doors and passages,

		\item \textbf{zap}: zap a wand
	\end{itemize}

	\textbf{For task which name consists "corridor":}\
Your goal is to explore the level
	and reach the stairs down.

	\textbf{For task which name consists “quest”:}\
Your goal is to explore the
	level, fight monsters, and navigate rooms and mazes to ultimately reach the
	stairs down.

	\textbf{For task which name consists “boxoban”:}\
You are playing Boxoban, a
	box-pushing game inspired by Sokoban. Your goal is to push the boulders onto the
	fountains on the map. You can push the boulders by walking into them, as long as
	there are no obstacles behind them.

	\textbf{For task which name consists “mazewalk”:}\
Your goal is to explore the
	level and reach the stairs down.

	\textbf{Otherwise:}\
Your goal is to get as far as possible in the game.

	In a moment I will present a history of actions and observations from the game.

	\textbf{Tip:} there is no point in outputting the same action over and over if
	nothing changes.

	\textbf{Additional Feedback (if provided):}\
I will also present lists of positive
	and negative feedback based on your previous attempts on this task, use this
	information to change your strategy and improve your performance. \
\texttt{Positive:
	{positive}}\
\texttt{Negative: {negative}}

	\textbf{PLAY!}
\end{MyGreenBox}

\begin{MyGreenBox}
	[frametitle={\textbf{Iteration 1 MiniHack Prompt}}] You are an agent playing MiniHack.
	The following are the possible actions you can take in the game, followed by a
	short description of each action:

	\begin{itemize}
		\item \textbf{north}: move north,

		\item \textbf{east}: move east,

		\item \textbf{south}: move south,

		\item \textbf{west}: move west,

		\item \textbf{northeast}: move northeast,

		\item \textbf{southeast}: move southeast,

		\item \textbf{southwest}: move southwest,

		\item \textbf{northwest}: move northwest,

		\item \textbf{far north}: move far north,

		\item \textbf{far east}: move far east,

		\item \textbf{far south}: move far south,

		\item \textbf{far west}: move far west,

		\item \textbf{far northeast}: move far northeast,

		\item \textbf{far southeast}: move far southeast,

		\item \textbf{far southwest}: move far southwest,

		\item \textbf{far northwest}: move far northwest,

		\item \textbf{up}: go up the stairs,

		\item \textbf{down}: go down the stairs,

		\item \textbf{wait}: rest one move while doing nothing,

		\item \textbf{more}: display more of the message,

		\item \textbf{apply}: apply (use) a tool,

		\item \textbf{close}: close an adjacent door,

		\item \textbf{open}: open an adjacent door,

		\item \textbf{eat}: eat something,

		\item \textbf{force}: force a lock,

		\item \textbf{kick}: kick an enemy or a locked door or chest,

		\item \textbf{loot}: loot a box on the floor,

		\item \textbf{pickup}: pick up things at the current location if there are any,

		\item \textbf{pray}: pray to the gods for help,

		\item \textbf{puton}: put on an accessory,

		\item \textbf{quaff}: quaff (drink) something,

		\item \textbf{search}: search for hidden doors and passages,

		\item \textbf{zap}: zap a wand
	\end{itemize}

	\textbf{For task name contains “corridor”:}\\ Your goal is to explore the
	level and reach the stairs down.

	\textbf{Notes:}
	\begin{enumerate}
		\item The target stairs down is always at the other room on the map, so you
			should prioritize exploring the other room.

		\item Whenever you encounter a monster (giant rat) in a room, you should
			always try to run to the direction of the corridor.

		\item Whenever you encounter a monster (giant rat) on the corridor, you
			should always try to kill the monster (giant rat).

		\item Once you reach the second room on the map, you should explore the room
			fully to find the target stairs down, but whenever you encounter a monster
			(giant rat), you should always try to run to the direction of the corridor.
	\end{enumerate}

	\textbf{Example:} \begin{verbatim}
                        ---
                        .<|
                        ..@#
                         .|
\end{verbatim}

	\textbf{For task name contains “quest”:}\\ Your goal is to explore the level, fight
	monsters, and navigate rooms and mazes to ultimately reach the stairs down.

	\textbf{Notes:}
	\begin{enumerate}
		\item You should never cross (lava) without necessary ability or equipment.

		\item Always explore the area approachable (no need to cross the lava) well enough
			to find sufficient support. Below is an example:
	\end{enumerate}

	\textbf{Map Example:} \begin{verbatim}
                        --------------
                        |@....}......---
                        |.(...}.......F-
                        |.....}.....................|
                        |.....}........--------     |
                        |.....}......---
                        --------------
\end{verbatim}

	\begin{enumerate}
		\setcounter{enumi}{2}

		\item You should always try to apply the tool you can get, even though you
			do not know the effect.

		\item You should always try to apply your role's ability, even though you do
			not know the effect.
	\end{enumerate}

	\textbf{For task name contains “boxoban”:}\\ You are playing Boxoban, a box-pushing
	game inspired by Sokoban. Your goal is to push the boulders onto the fountains
	on the map. You can push the boulders by walking into them, as long as there are
	no obstacles behind them.

	\textbf{Notes:} For this task, you should follow the below rules with priority
	from top to down:
	\begin{enumerate}
		\item You should remember the locations of the fountains in the beginning.

		\item You should never push a boulder in a direction if in this direction,
			the next spot next to the boulder is not empty or is not the fountain.

		\item You should never push a boulder to any L-shaped wall configuration (two
			adjacent walls).
	\end{enumerate}

	\textbf{Example (dead corner):}
	\begin{verbatim}
                                  ##########
                                  #`.....###
\end{verbatim}

	\begin{enumerate}
		\setcounter{enumi}{3}

		\item You should never push a boulder next to a wall unless, in the
			direction perpendicular to the line joining the boulder and its neighbor
			wall, there is at least one fountain left open such that between this fountain
			and the boulder all spots are empty.

		\item You should never push a boulder if it is already on the fountain.

		\item If a boulder and a fountain are on the same row or column, and all
			spots between this boulder and this fountain are empty, and the first spot
			next to the boulder on the opposite direction of which the fountain locates
			is empty, you should always push this boulder to the corresponding
			fountain.
	\end{enumerate}

	\textbf{Map Example (established path):} \begin{verbatim}
                                  ##########
                                  #......<##
                                  #.#@.{`.##
                                  #.``##.#.#
                                  #..`...{.#
                                  #.#.{....#
                                  #######.{#
                                  #######..#
                                  #######..#
                                  ##########
\end{verbatim}

	\textbf{Example (dead corner reminder):} \begin{verbatim}
                                  ##########
                                  #`.....###
\end{verbatim}

	\textbf{For task name contains “mazewalk”:}\\ Your goal is to explore the
	level and reach the stairs down.

	\textbf{Notes:} For this task, your action space will be limited to: \begin{verbatim}
{"north": "move north",
 "east":  "move east",
 "south": "move south",
 "west":  "move west"}
\end{verbatim}

	You will not choose any actions other than these four. Follow strictly:
	\begin{enumerate}
		\item If the block on your east is empty, move east.

		\item If east is blocked and north is open, move north.

		\item If east and north are both blocked, and west is open, move west.

		\item If east, north, and west are blocked, move south.
	\end{enumerate}

	\textbf{Otherwise:}\\ Your goal is to get as far as possible in the game.

	In a moment I will present a history of actions and observations from the game.

	\textbf{Tip:} there is no point in outputting the same action over and over if
	anything changes.

	\textbf{PLAY!}
\end{MyGreenBox}

\begin{MyGreenBox}
	[frametitle={\textbf{Iteration 2 MiniHack Prompt}}] You are an agent playing MiniHack.
	The following are the possible actions you can take in the game:

	\begin{itemize}
		\item \textbf{north}: move north,

		\item \textbf{east}: move east,

		\item \textbf{south}: move south,

		\item \textbf{west}: move west,

		\item \textbf{northeast}: move northeast,

		\item \textbf{southeast}: move southeast,

		\item \textbf{southwest}: move southwest,

		\item \textbf{northwest}: move northwest,

		\item \textbf{far north}: move far north,

		\item \textbf{far east}: move far east,

		\item \textbf{far south}: move far south,

		\item \textbf{far west}: move far west,

		\item \textbf{far northeast}: move far northeast,

		\item \textbf{far southeast}: move far southeast,

		\item \textbf{far southwest}: move far southwest,

		\item \textbf{far northwest}: move far northwest,

		\item \textbf{up}: go up the stairs,

		\item \textbf{down}: go down the stairs,

		\item \textbf{wait}: rest one move while doing nothing,

		\item \textbf{more}: display more of the message,

		\item \textbf{apply}: apply (use) a tool,

		\item \textbf{close}: close an adjacent door,

		\item \textbf{open}: open an adjacent door,

		\item \textbf{eat}: eat something,

		\item \textbf{force}: force a lock,

		\item \textbf{kick}: kick an enemy or a locked door or chest,

		\item \textbf{loot}: loot a box on the floor,

		\item \textbf{pickup}: pick up things at the current location if there are any,

		\item \textbf{pray}: pray to the gods for help,

		\item \textbf{puton}: put on an accessory,

		\item \textbf{quaff}: quaff (drink) something,

		\item \textbf{search}: search for hidden doors and passages,

		\item \textbf{zap}: zap a wand
	\end{itemize}

	\textbf{For task name contains “corridor”:}\\ \textbf{Your goal:} Your goal is
	to explore the level and reach the stairs down.

	\textbf{Notes:}

	Most important notice before you start: "." in the map means room, so if your
	current space is marked by "." you are in the room. In the map, "\#" means
	corridor. You should never misrecognize the type of location you are currently
	at as this is crucial for your game. Always be certain about your current
	location before you decide an action.

	\textit{You should permanently treat “<” the same as “.”}

	\begin{enumerate}
		\item The target stairs down is always at the other room on the map, so you
			should prioritize exploring the other room. Remember, the goal spot is
			always at the far east of the map.

		\item Whenever you haven’t moved onto the corridor (“\#”) yet, find one “\#”
			as soon as possible and move onto it.
	\end{enumerate}

	\textbf{Example:}
	\begin{verbatim}
                        ---
                        .<|
                        ..@#
                         .|
\end{verbatim}

	\textbf{Example:}
	\begin{verbatim}
                      -----
                      |..<|
                      |@...
                      |...|
                      -----
\end{verbatim}

	\begin{enumerate}
		\setcounter{enumi}{2}

		\item Whenever you encounter a monster (giant rat) in a room, you should
			always try to run to the direction of the corridor. As soon as you reach the
			second block on the corridor, you should stop running away after moving
			west for a max of 3 steps and start fighting the rat. Fight until there are
			no rats on your adjacent spots.
	\end{enumerate}

	\textbf{Example:} \begin{verbatim}
                       ----       ---
                       ...|       |r...%
                      |....#######@r....
                      |<..|       |r....
                      ----          ---
\end{verbatim}

	You should only fight the rat when there is exactly one rat adjacent (in the 8
	surrounding spots).

	\begin{enumerate}
		\setcounter{enumi}{3}

		\item Whenever you encounter a monster (giant rat) on the corridor, you
			should always try to kill the monster immediately. Never retreat back into
			the room.
	\end{enumerate}

	\textbf{Example:}
	\begin{verbatim}
                      -----
                      |...|
                      |....##%##@r.
                      |..<|
                       ----
\end{verbatim}

	\textbf{Example:}
	\begin{verbatim}
                      -----
                      |...|
                      |....@r##%##.
                      |..<|
                       ----
\end{verbatim}

	\begin{enumerate}
		\setcounter{enumi}{4}

		\item Once you reach the second room on the map, you should explore the room
			fully to find the target stairs down, but whenever you encounter a monster,
			you should always try to run to the direction of the corridor. Once in the
			second room, never return to the corridor or first room unless actively
			fleeing a monster.
	\end{enumerate}

	\textbf{Example:} \begin{verbatim}
                      ----
                      |...|       |..............
                      |<.%.#%#####............@..
                      |...|       |..............
                      ----
\end{verbatim}

	\begin{enumerate}
		\setcounter{enumi}{5}

		\item In case you return to the first room (where you can see “<”),
			immediately find the corridor “\#” again and go east as much as possible.
	\end{enumerate}

	\textbf{For task name contains “quest”:}\\ \textbf{Your goal:} Your goal is to
	explore the level, fight monsters, and navigate rooms and mazes to ultimately
	reach the stairs down.

	\textbf{Notes:}
	\begin{enumerate}
		\item You should never cross \}(lava) without necessary ability or equipment.

		\item Always explore approachable areas (no need to cross the lava) well enough
			to find sufficient support. Below is an example:
	\end{enumerate}

	\textbf{Map Example:} \begin{verbatim}
                        --------------
                        |@....}......---
                        |.(...}.......F-
                        |.....}.....................|
                        |.....}........--------     |
                        |.....}......---
                        --------------
\end{verbatim}

	\begin{enumerate}
		\setcounter{enumi}{2}

		\item You should always try to apply the tool you can get, even if you do
			not know its effect.

		\item You should always try to apply your role’s ability, even if you do not
			know its effect.
	\end{enumerate}

	\textbf{For task name contains “boxoban”:}\\ \textbf{Your goal:} You are
	playing Boxoban, a box-pushing game inspired by Sokoban. Your goal is to push
	the boulders onto the fountains on the map. You can push the boulders by walking
	into them, as long as there are no obstacles behind them.

	\textbf{Notes:} For this task, follow these rules in priority order:
	\begin{enumerate}
		\item Remember the locations of the fountains at the start.

		\item Never push a boulder if the next spot is not empty or not a fountain.

		\item Never push a boulder into an L-shaped wall configuration (two adjacent
			walls).
	\end{enumerate}

	\textbf{Example (dead corner):} \begin{verbatim}
                                  ##########
                                  #`.....###
\end{verbatim}

	\textbf{Visual Deadlock Examples (NEVER DO THESE):}

	\textbf{Case 1:} Before Push:
	\begin{verbatim}
                                  ##########
                                  ##########
                                  #<########
                                  #.########
                                  #.########
                                  #..{.#####
                                  #@.``#####
                                  #`{.`{.###
                                  #......###
                                  ##########
\end{verbatim}
	Action: Push South After Push:
	\begin{verbatim}
                                  ##########
                                  ##########
                                  #<########
                                  #.########
                                  #.########
                                  #..{.#####
                                  #..``#####
                                  #@{.`{.###
                                  #`.....###
                                  ##########
\end{verbatim}
	Result: Boulder stuck – can’t move due to walls.

	\textbf{Case 2:} Before Push: \begin{verbatim}
                                  ##########
                                  ####...{`#
                                  #..#.`@..#
                                  #.{.#.#.<#
                                  #.``..##.#
                                  ##...{##.#
                                  ########.#
                                  #####...{#
                                  ######...#
                                  ##########
\end{verbatim}
	Action: Push WEST After Push: \begin{verbatim}
                                  ##########
                                  ####...{`#
                                  #..#`@...#
                                  #.{.#.#.<#
                                  #.``..##.#
                                  ##...{##.#
                                  ########.#
                                  #####...{#
                                  ######...#
                                  ##########
\end{verbatim}
	Result: Can’t move boulder any more.

	\textbf{More Thorough Examples:}

	\textbf{Example 1 – Dead Corner:} Before:
	\begin{verbatim}
                                  ##########
                                  ####...{`#
                                  #..#.`@..#
                                  #.{.#.#.<#
                                  #.``..##.#
                                  ##...{##.#
                                  ########.#
                                  #####...{#
                                  ######...#
                                  ##########
\end{verbatim}
	After pushing west: \begin{verbatim}
                                  ##########
                                  ####...{`#
                                  #..#`@...#
                                  #.{.#.#.<#
                                  #.``..##.#
                                  ##...{##.#
                                  ########.#
                                  #####...{#
                                  ######...#
                                  ##########
\end{verbatim}

	\textbf{Example 2 – Dead Corner:} Before:
	\begin{verbatim}
                                  ##########
                                  ##########
                                  #<########
                                  #.########
                                  #.########
                                  #..{.#####
                                  #@.``#####
                                  #`{.`{.###
                                  #......###
                                  ##########
\end{verbatim}
	After pushing south: \begin{verbatim}
                                  ##########
                                  ##########
                                  #<########
                                  #.########
                                  #.########
                                  #..{.#####
                                  #..``#####
                                  #@{.`{.###
                                  #`.....###
                                  ##########
\end{verbatim}

	\textbf{Example 3 – Dead Corner:} Before:
	\begin{verbatim}
                                  ##########
                                  ####...{`#
                                  #..#`....#
                                  #.@.#.#.<#
                                  #.``..##.#
                                  ##...{##.#
                                  ########.#
                                  #####...{#
                                  ######...#
                                  ##########
\end{verbatim}
	After pushing north: \begin{verbatim}
                                  ##########
                                  ####...{`#
                                  #..#`....#
                                  #...#.#.<#
                                  #.@`..##.#
                                  ##`..{##.#
                                  ########.#
                                  #####...{#
                                  ######...#
                                  ##########
\end{verbatim}

	\textbf{Example 4 – Dead Corner:} Before:
	\begin{verbatim}
                                  ##########
                                  ####...{`#
                                  #..{`....#
                                  #.{.#.#.<#
                                  #..`..##.#
                                  ##.@`.##.#
                                  ########.#
                                  #####...{#
                                  ######...#
                                  ##########
\end{verbatim}
	After pushing east: \begin{verbatim}
                                  ##########
                                  ####...{`#
                                  #..{`....#
                                  #.{.#.#.<#
                                  #..`..##.#
                                  ##..@`##.#
                                  ########.#
                                  #####...{#
                                  ######...#
                                  ##########
\end{verbatim}

	\textbf{Terminal State Check:} Remember all fountain locations initially.
	After each push, if a fountain spot “{” is covered by a boulder, mark that spot terminated.

	\textbf{Example 1 – Termination:} Before: \begin{verbatim}
                                  ##########
                                  ####...{`#
                                  #..#`....#
                                  #.{.#.#.<#
                                  #.``..##.#
                                  ##@..{##.#
                                  ########.#
                                  #####...{#
                                  ######...#
                                  ##########
\end{verbatim}

	After pushing north:

	\begin{verbatim}
                                  ##########
                                  ####...{`#
                                  #..#`....#
                                  #.`.#.#.<#
                                  #.@`..##.#
                                  ##...{##.#
                                  ########.#
                                  #####...{#
                                  ######...#
                                  ##########
\end{verbatim}

	\textbf{Example 2 – Termination:} Before:

	\begin{verbatim}
                                  ##########
                                  ####...{`#
                                  #..{`@...#
                                  #...#.#.<#
                                  #.``..##.#
                                  ##...{##.#
                                  ########.#
                                  #####...{#
                                  ######...#
                                  ##########
\end{verbatim}

	After pushing west:

	\begin{verbatim}
                                  ##########
                                  ####...{`#
                                  #..`@....#
                                  #...#.#.<#
                                  #.``..##.#
                                  ##...{##.#
                                  ########.#
                                  #####...{#
                                  ######...#
                                  ##########
\end{verbatim}

	4. You should never push a boulder next to a wall unless, in the perpendicular direction there remains an open fountain with all intermediate spots empty. 5. You should never push a boulder if it is already on a fountain. 6. If a boulder and a fountain are aligned in the same row or column, all intermediate spots empty, and the spot behind the boulder (opposite the fountain) is empty, push toward the fountain.

	\textbf{Map Example (established path):} \begin{verbatim}
                                  ##########
                                  #......<##
                                  #.#@.{`.##
                                  #.``##.#.#
                                  #..`...{.#
                                  #.#.{....#
                                  #######.{#
                                  #######..#
                                  #######..#
                                  ##########
\end{verbatim}

	7. Always explore to create more paths. 8. If a boulder is in a dead corner initially, do not waste steps pushing it.

	\textbf{Example (dead corner reminder):} \begin{verbatim}
                                  ##########
                                  #`.....###
\end{verbatim}

	\textbf{Below are two quick dead-corner examples:}

	\textbf{Example 1:} \begin{verbatim}
                                  ##########
                                  ####...{`#
                                  #..#`@...#  <-- Boulder in dead corner
                                  #.{.#.#.<#
                                  #.``..##.#
                                  ##...{##.#
                                  ########.#
                                  #####...{#
                                  ######...#
                                  ##########
\end{verbatim}

	\textbf{Example 2:} \begin{verbatim}
                                  ##########
                                  ##########
                                  #<########
                                  #.########
                                  #.########
                                  #..{.#####
                                  #..``#####
                                  #@{.`{.###
                                  #`.....###  <-- Boulder in dead corner
\end{verbatim}

	\textbf{Finally, a successful trajectory:}

	\textbf{Example 1 – Beginning:} \begin{verbatim}
                                  ##########
                                  #...{..{.#
                                  ##.`.{.`.#
                                  ####.#####
                                  ####....{#
                                  ###.`.`..#
                                  ###.@.#..#
                                  ####<#####
                                  ##########
                                  ##########
\end{verbatim}

	\textbf{move east:} \begin{verbatim}
                                  ##########
                                  #...{..{.#
                                  ##.`.{.`.#
                                  ####.#####
                                  ####....{#
                                  ###.`.`..#
                                  ###..@#..#
                                  ####<#####
                                  ##########
                                  ##########
\end{verbatim}

	\textbf{move north:} \begin{verbatim}
                                  ##########
                                  #...{..{.#
                                  ##.`.{.`.#
                                  ####.#####
                                  ####....{#
                                  ###.`@`..#
                                  ###...#..#
                                  ####<#####
                                  ##########
                                  ##########
\end{verbatim}

	\textbf{move east:} \begin{verbatim}
                                  ##########
                                  #...{..{.#
                                  ##.`.{.`.#
                                  ####.#####
                                  ####....{#
                                  ###.`.@`.#
                                  ###...#..#
                                  ####<#####
                                  ##########
                                  ##########
\end{verbatim}

	\textbf{move east:} \begin{verbatim}
                                  ##########
                                  #...{..{.#
                                  ##.`.{.`.#
                                  ####.#####
                                  ####....{#
                                  ###.`..@`#
                                  ###...#..#
                                  ####<#####
                                  ##########
                                  ##########
\end{verbatim}

	\textbf{move south:} \begin{verbatim}
                                  ##########
                                  #...{..{.#
                                  ##.`.{.`.#
                                  ####.#####
                                  ####....{#
                                  ###.`...`#
                                  ###...#@.#
                                  ####<#####
                                  ##########
                                  ##########
\end{verbatim}

	\textbf{move east:} \begin{verbatim}
                                  ##########
                                  #...{..{.#
                                  ##.`.{.`.#
                                  ####.#####
                                  ####....{#  <----target this fountain
                                  ###.`...`#
                                  ###...#.@#
                                  ####<#####
                                  ##########
                                  ##########
\end{verbatim}

	\textbf{move north:} \begin{verbatim}
                                  ##########
                                  #...{..{.#
                                  ##.`.{.`.#
                                  ####.#####
                                  ####....`#  <----boulder now on fountain
                                  ###.`...@#
                                  ###...#..#
                                  ####<#####
                                  ##########
                                  ##########
\end{verbatim}

	\textbf{move west:} \begin{verbatim}
                                  ##########
                                  #...{..{.#
                                  ##.`.{.`.#
                                  ####.#####
                                  ####....`#
                                  ###.`..@.#
                                  ###...#..#
                                  ####<#####
                                  ##########
                                  ##########
\end{verbatim}

	\textbf{move west:} \begin{verbatim}
                                  ##########
                                  #...{..{.#
                                  ##.`.{.`.#
                                  ####.#####
                                  ####....`#
                                  ###.`.@..#
                                  ###...#..#
                                  ####<#####
                                  ##########
                                  ##########
\end{verbatim}

	\textbf{move west:} \begin{verbatim}
                                  ##########
                                  #...{..{.#
                                  ##.`.{.`.#
                                  ####.#####
                                  ####....`#
                                  ###.`@...#
                                  ###...#..#
                                  ####<#####
                                  ##########
                                  ##########
\end{verbatim}

	\textbf{move south:} \begin{verbatim}
                                  ##########
                                  #...{..{.#
                                  ##.`.{.`.#
                                  ####.#####
                                  ####....`#
                                  ###.`....#
                                  ###..@#..#
                                  ####<#####
                                  ##########
                                  ##########
\end{verbatim}

	\textbf{move west:} \begin{verbatim}
                                  ##########
                                  #...{..{.#
                                  ##.`.{.`.#
                                  ####.#####
                                  ####....`#
                                  ###.`....#
                                  ###.@.#..#
                                  ####<#####
                                  ##########
                                  ##########
\end{verbatim}

	\textbf{move north:} \begin{verbatim}
                                  ##########
                                  #...{..{.#
                                  ##.`.{.`.#
                                  ####.#####
                                  ####`...`#
                                  ###.@....#
                                  ###...#..#
                                  ####<#####
                                  ##########
                                  ##########
\end{verbatim}

	\textbf{move north:} \begin{verbatim}
                                  ##########
                                  #...{..{.#
                                  ##.`.{.`.#
                                  ####`#####
                                  ####@...`#
                                  ###......#
                                  ###...#..#
                                  ####<#####
                                  ##########
                                  ##########
\end{verbatim}

	\textbf{move north:} \begin{verbatim}
                                  ##########
                                  #...{..{.#
                                  ##.``{.`.#
                                  ####@#####
                                  ####....`#
                                  ###......#
                                  ###...#..#
                                  ####<#####
                                  ##########
                                  ##########
\end{verbatim}

	\textbf{move north:} \begin{verbatim}
                                  ##########
                                  #...`..{.#
                                  ##.`@{.`.#
                                  ####.#####
                                  ####....`#
                                  ###......#
                                  ###...#..#
                                  ####<#####
                                  ##########
                                  ##########
\end{verbatim}

	\textbf{move east:} \begin{verbatim}
                                  ##########
                                  #...`..{.#
                                  ##.`.@.`.#
                                  ####.#####
                                  ####....`#
                                  ###......#
                                  ###...#..#
                                  ####<#####
                                  ##########
                                  ##########
\end{verbatim}

	\textbf{move north:} \begin{verbatim}
                                  ##########
                                  #...`@.{.#
                                  ##.`.{.`.#
                                  ####.#####
                                  ####....`#
                                  ###......#
                                  ###...#..#
                                  ####<#####
                                  ##########
                                  ##########
\end{verbatim}

	\textbf{move east:} \begin{verbatim}
                                  ##########
                                  #...`.@{.#
                                  ##.`.{.`.#
                                  ####.#####
                                  ####....`#
                                  ###......#
                                  ###...#..#
                                  ####<#####
                                  ##########
                                  ##########
\end{verbatim}

	\textbf{move east:} \begin{verbatim}
                                  ##########
                                  #...`..@.#
                                  ##.`.{.`.#
                                  ####.#####
                                  ####....`#
                                  ###......#
                                  ###...#..#
                                  ####<#####
                                  ##########
                                  ##########
\end{verbatim}

	\textbf{move east:} \begin{verbatim}
                                  ##########
                                  #...`..{@#
                                  ##.`.{.`.#
                                  ####.#####
                                  ####....`#
                                  ###......#
                                  ###...#..#
                                  ####<#####
                                  ##########
                                  ##########
\end{verbatim}

	\textbf{move south:} \begin{verbatim}
                                  ##########
                                  #...`..{.#
                                  ##.`.{.`@#
                                  ####.#####
                                  ####....`#
                                  ###......#
                                  ###...#..#
                                  ####<#####
                                  ##########
                                  ##########
\end{verbatim}

	\textbf{move west:} \begin{verbatim}
                                  ##########
                                  #...`..{.#
                                  ##.`.{`@.#
                                  ####.#####
                                  ####....`#
                                  ###......#
                                  ###...#..#
                                  ####<#####
                                  ##########
                                  ##########
\end{verbatim}

	\textbf{move west:} \begin{verbatim}
                                  ##########
                                  #...`..{.#
                                  ##.`.`@..#  <--- now boulder on fountain
                                  ####.#####
                                  ####....`#
                                  ###......#
                                  ###...#..#
                                  ####<#####
                                  ##########
                                  ##########
\end{verbatim}

	\textbf{Now we have successfully pushed three boulders to the fountains.}

	\textbf{For task name contains “mazewalk”:}\\ \textbf{Your goal:} Your goal is to explore the level and reach the stairs down.

	\textbf{Notes:} For this task, your action space is limited to: \begin{verbatim}
{"north": "move north",
 "east":  "move east",
 "south": "move south",
 "west":  "move west"}
\end{verbatim}

	You must only choose these four directions:

	\begin{enumerate}\item If the block on your east is empty, move east.

	\item If east is blocked and north is open, move north.

	\item If east and north are both blocked, and west is open, move west.

	\item If east, north, and west are blocked, move south.\end{enumerate}

	\textbf{Absolute Wall-Following Protocol:}\\ \textit{Core Principle: “Follow priority order, probe dark directions ONLY when they're current priority.”}\\ \textit{Movement Validation:}\\ – Priority rotation ONLY occurs AFTER SUCCESSFUL MOVEMENT.\\ – Wall collisions preserve current priority order.

	\begin{enumerate}\item Initialization: CURRENT\_PRIORITY = [EAST, NORTH, WEST, SOUTH]; permanent\_walls = \{\}. Move east until first wall.

	\item At each position, check directions in CURRENT\_PRIORITY: \begin{itemize}\item If visible (“.” or “<”), move immediately and rotate per rules.

	\item If dark, attempt move: \begin{itemize}\item Success: continue protocol.

	\item Wall collision: mark permanent\_walls, skip direction, continue.\end{itemize}\end{itemize}

	\item Rotation rules: \begin{enumerate}\item 1st priority move: rotate clockwise (e.g. ENWS → SENW).

	\item 2nd priority: no rotation.

	\item 3rd priority: rotate counter-clockwise (ENWS → NWSE).

	\item 4th priority: reverse (swap 1$\leftrightarrow$3, 2$\leftrightarrow$4).\end{enumerate}

	\item Terminal Condition: immediately go to stairs down (">") when visible; never treat "<" specially.\end{enumerate}

	\textbf{Walkthrough Example:}\\ Initial: [E,N,W,S]\\ 1. Move E until wall. 2. At start: E (wall) → N (open) → move north (no rotation). 3. Next: E (open) → move east (rotate to SENW).\\

	\textbf{Otherwise:}\\ \textbf{Your goal:} Your goal is to get as far as possible in the game.

	In a moment I will present a history of actions and observations from the game.

	\textbf{Tip:} there is no point in outputting the same action over and over if nothing changes.

	\textbf{PLAY!} }
\end{MyGreenBox}
    \newpage

    \section{Qualitative Observations of MiniHack Agent Performance}
    \label{appendix:minihack_obs}

\centering
\begin{table}
	\begin{tabular}{cc>{\raggedright\arraybackslash}p{5.5cm}}
		\toprule \multicolumn{1}{c}{\textbf{}}                                           & \multicolumn{1}{c}{\textbf{Turn}} & \multicolumn{1}{c}{\textbf{Description}}         \\
		\midrule \rowcolor{gray!10} \includegraphics[scale=0.07]{figs/emojis/mini_1.png} & 0                                 & Target Navigation Effectiveness                  \\
		\midrule \rowcolor{gray!10} \includegraphics[scale=0.07]{figs/emojis/mini_2.png} & 0                                 & Efficient Exploration and Map Memory Utilization \\
		\midrule \rowcolor{gray!10} \includegraphics[scale=0.07]{figs/emojis/mini_3.png} & 0                                 & Hazard Awareness and Equipment Utilization       \\
		\midrule \rowcolor{gray!10} \includegraphics[scale=0.07]{figs/emojis/mini_4.png} & 0                                 & Boulder Manipulation Strategy                    \\
		\midrule \rowcolor{gray!10} \includegraphics[scale=0.07]{figs/emojis/mini_5.png} & 0                                 & Combat Engagement and Survival                   \\
		\midrule \rowcolor{gray!10} \includegraphics[scale=0.07]{figs/emojis/mini_6.png} & 0                                 & Role-Specific Ability Utilization                \\
		\midrule \rowcolor{gray!30} \includegraphics[scale=0.07]{figs/emojis/mini_7.png} & 1                                 & Spatial Awareness and Interpretation             \\
		\midrule \rowcolor{gray!30} \includegraphics[scale=0.07]{figs/emojis/mini_8.png} & 1                                 & Object Pickup Efficiency                         \\
		\midrule \rowcolor{gray!60} \includegraphics[scale=0.07]{figs/emojis/mini_9.png} & 2                                 & Giant Rats Encounter Handling                    \\
		\bottomrule
	\end{tabular}
	\caption{\raggedright Metrics and Turn of Induction for MiniHack}
	\label{tab:metrics_mini}
\end{table}
\begin{flushleft}
	The induced metrics and the agent's per-task performance are shown in Table \ref{tab:metrics_mini}
	and Table \ref{tab:metric_mini_perf}, respectively. A substantial improvement in
	the agent's task completion performance is observed from \emph{Iteration 1} to
	\emph{Iteration 2}, with the agent achieving both a higher environment score
	and trajectory performance on the held-out tasks compared to the baseline
	agent. Specifically, the agent's performance on metrics increased
	correspondingly to code changes, like
	\includegraphics[scale=0.05]{figs/emojis/mini_1.png}
	improving from $16.67\%$ to $41.67\%$, demonstrating the utility of AutoLibra
	for fine-grained agent improvement. This is discussed in more detail in the following
	sections.

	\subsection{Extracted Metrics and Improvements}
	The metrics induced by AutoLibra (Table \ref{tab:metrics_mini}) capture the behavior
	of the agent through all iterations, with coverage of 83\% at \emph{Iteration
	0} to 88\% at \emph{Iteration 2}. Due to the diversity of tasks, metrics
	pertain to one or two tasks, matching the expectation that AutoLibra should generate
	fine-grained metrics. Notably, the metrics generated at \emph{Iteration 0} are
	nearly comprehensive as they demonstrate good coverage of all four MiniHack
	tasks, while the later metrics were found to specifically describe detailed behaviors
	for targeting one task each. Code changes were selected to specifically target
	improvements in given metrics, and as seen in Table \ref{tab:metric_mini_perf},
	the agent's performance on the targeted metric improved significantly in the
	iteration following the code change. This demonstrates the utility of AutoLibra
	for fine-grained agent improvement, as well as the human interpretability of
	the induced metrics. On the other hand, some metrics remain unchanged over
	iterations. The unchanged metrics are
	\includegraphics[scale=0.05]{figs/emojis/mini_3.png}
	,
	\includegraphics[scale=0.05]{figs/emojis/mini_4.png}
	,
	\includegraphics[scale=0.05]{figs/emojis/mini_6.png}
	, and
	\includegraphics[scale=0.05]{figs/emojis/mini_8.png}
	. These metrics are all tightly related to Boxoban or Quest environments. As
	these two environments contain high level of randomness, interactions with system,
	and sophisticated planning, our prompt and example based guidance do not provide
	obvious improvements. This result also indicates that simply coding might not
	be sufficient if the gap between the agent's capability and the task's
	complexity is too significant.

	\textbf{Iteration 0} The agent's behavior is stochastic and repetitive, with
	no utilization of memory, planning, and goal awareness. Although the agent has
	reasoning before taking action, the decided action shows no correlation with
	the goal or the environment. \texttt{Target Navigation Effectiveness}
	\includegraphics[scale=0.05]{figs/emojis/mini_1.png}
	was identified as the core metric that evaluates the performance of all tasks
	except Boxoban. As all three other tasks require the agent to explore the target
	exit stairs,
	\includegraphics[scale=0.05]{figs/emojis/mini_1.png}
	serves as the fundamental test of the agent's goal awareness for these tasks. Since
	Boxoban has a different winning condition, \texttt{Boulder Manipulation
	Strategy}
	\includegraphics[scale=0.05]{figs/emojis/mini_4.png}
	induced in this iteration targets describing the overall goal awareness for the
	agent in Boxoban environment. The remaining four metrics each cover two to
	three tasks on a more detailed level.

	Based on the metrics induced in \emph{Iteration 0}, several changes to the
	agent code were implemented:
	\begin{itemize}
		\item Augmentation of subtask-level behavioral restriction guidance with the
			single-shot example, targeting improvement of \texttt{Boulder Manipulation
			Strategy}
			\includegraphics[scale=0.05]{figs/emojis/mini_4.png}
			, \texttt{Combat Engagement and Survival}
			\includegraphics[scale=0.05]{figs/emojis/mini_5.png}
			, and \texttt{Hazard Awareness and Equipment Utilization}
			\includegraphics[scale=0.05]{figs/emojis/mini_3.png}
			.

		\item Meta-prompting by providing subtask instructions targeting improvement
			of \texttt{Role-Specific Ability Utilization}
			\includegraphics[scale=0.05]{figs/emojis/mini_6.png}
			.

		\item Augmentation of subtask-level winning strategy with few-shot example,
			targeting improvement of \texttt{Target Navigation Effectiveness}
			\includegraphics[scale=0.05]{figs/emojis/mini_1.png}
			and \texttt{Efficient Exploration and Map Memory Utilization}
			\includegraphics[scale=0.05]{figs/emojis/mini_2.png}
			.
	\end{itemize}

	\textbf{Iteration 1} The changes made after \emph{Iteration 0} did not yield
	substantial improvements in agent task completion performance, but behavior changes
	matching the metrics were observed and insights on prompt improvements were obtained.
	A decrease in
	\includegraphics[scale=0.05]{figs/emojis/mini_1.png}
	performance of 8.3\% was observed between \emph{Iteration 0} and \emph{Iteration
	1}, indicating that the code changes confuse the agent which reduces their
	capability of goal recognition, and the same reduction is also observed for
	\includegraphics[scale=0.05]{figs/emojis/mini_2.png}
	. Three causes of these reductions are recognized. Firstly, the overly simplified
	strategy for MazeWalk provides no positive intuitions for the agent when loops
	exist in the map. Secondly, the seemingly straightforward behavioral restrictions
	with examples do not appear sufficient enough for the agent to avoid false action
	as the agent failed to comprehend the map thorough enough. Thirdly, the randomness
	of Quest is so high that existing guidance fails to cover new randomly generated
	situations. Based on \emph{Iteration 1's} result, MazeWalk and Corridor Fight
	both appear solvable given a more optimal strategy, and Quest and Boxoban,
	given their high randomness and complexity, do not appear easily solvable, so
	providing more examples seem to be the only direction of improvement. Moreover,
	two new metrics are induced, one focusing on Corridor Fight and the other focusing
	on Quest.

	Based on the metrics induced in \emph{Iteration 1}, new changes are listed below:
	\begin{itemize}
		\item Augmentation of existing subtask-level advanced strategy with step-by-step
			decision-making instructions and few-shot examples, targeting improvement
			of \texttt{Target Navigation Effectiveness}
			\includegraphics[scale=0.05]{figs/emojis/mini_1.png}
			and \texttt{Efficient Exploration and Map Memory Utilization}
			\includegraphics[scale=0.05]{figs/emojis/mini_2.png}
			.

		\item Augmentation of existing subtask-level strategy with corner cases handling
			guidance and few-shot examples, targeting improvement of \texttt{Combat
			Engagement and Survival}
			\includegraphics[scale=0.05]{figs/emojis/mini_5.png}
			, \texttt{Hazard Awareness and Equipment Utilization}
			\includegraphics[scale=0.05]{figs/emojis/mini_3.png}
			, and \texttt{Spatial Awareness and Interpretation}
			\includegraphics[scale=0.05]{figs/emojis/mini_7.png}
			.

		\item Augmentation of existing few-shot examples with a full trajectory of the
			complete winning task, targeting improvement of \texttt{Boulder
			Manipulation Strategy}
			\includegraphics[scale=0.05]{figs/emojis/mini_4.png}
			.

		\item Meta-prompting by providing more in-depth subtask instructions,
			targeting improvement of \texttt{Role-Specific Ability Utilization}
			\includegraphics[scale=0.05]{figs/emojis/mini_6.png}
			and \texttt{Object Pickup Efficiency}
			\includegraphics[scale=0.05]{figs/emojis/mini_8.png}
			.
	\end{itemize}

	\textbf{Iteration 2} We observe a significant increase in
	\includegraphics[scale=0.05]{figs/emojis/mini_1.png}
	and
	\includegraphics[scale=0.05]{figs/emojis/mini_2.png}
	indicating the code changes successfully improve agent's ability in goal
	awareness and goal reaching efficiency. Based on task-level observation, the
	agent succeeded in most MazeWalk and Corridor Fight tasks. The large increase in
	\includegraphics[scale=0.05]{figs/emojis/mini_5.png}
	and
	\includegraphics[scale=0.05]{figs/emojis/mini_7.png}
	shows the agent's strong performance in the Corridor Fight task. More
	specifically, the agent now demonstrates both the capability of choosing optimal
	action based on its current location and the capability of strategically handling
	dangerous enemies—giant rats—in an emergency. The new induced metric also
	further proves the agent's rat-handling ability. For instance, when the agent
	encounters multiple rats at the same time, it will immediately retreat to the middle
	of the single-way corridor to engage rats in combat individually. By doing so,
	the agent can easily beat rats one by one and navigate to the goal without danger.
	However, as all the remaining metrics show no improvements, we also observe
	the agent's poor performance on Boxoban and Quest. This result shows that our current
	targeting improvement strategy of adding abundant few-shot examples does not seem
	effective given the gap between the agent's reasoning ability and the high randomness
	and complexity of Boxoban and Quest. For example, despite more than ten examples
	of pushing a boulder to a dead corner, a prohibited behavior has been
	presented, the agent will still randomly push a boulder to the L-shaped dead corner
	formed by two walls which causes this boulder to become immovable.

	\subsection{Held-Out Task Performance}
	For the held-out tasks, the agent task completion rate improved from \emph{Iteration
	0} to \emph{Iteration 2}, with scores from 0\% to 25\%. This result further
	proves that the improvements realized by AutoLibra are generalizable to unseen
	tasks. However, some failed cases also provide valuable insights into
	prospective future iterations. For instance, the complete set of Corridor
	tasks consists of scenarios where the target downstairs may be located west of
	the agent’s starting point, unlike Corridor Fight, where the target always
	lies to the east. Consequently, an instruction that rigidly directs the agent
	to explore only eastward fails to generalize to the broader Corridor tasks. This
	outcome underscores the importance of crafting instructions that are specific yet
	flexible: they should provide sufficiently detailed, high-level strategies without
	relying on overly prescriptive, task-specific steps. By focusing on generalizable
	principles rather than rigid directions, the agent is better equipped to adapt
	to a wider range of scenarios while still benefiting from structured guidance.
	The agent's performance on the held-out tasks is shown in Table \ref{tab:heldout_mini}.
\end{flushleft}

    \section{NNetNav-Live Induced Metrics}
    \label{appendix:nnetnav_live}
    \begin{wraptable}
    [19]{r}{0.60\textwidth} 
    \centering
    \small
    \vspace{-10pt}
    \begin{tabular}{ccl}
        \toprule

\multicolumn{1}{c}{It.} & \multicolumn{1}{l}{Metric}               \\
        \midrule \rowcolor{gray!10} 0     & Navigation Accuracy                      \\
        \midrule \rowcolor{gray!10} 0     & Search Term Accuracy                     \\
        \midrule \rowcolor{gray!10} 0     & Information Extraction Details           \\
        \midrule \rowcolor{gray!10} 0     & Task Goal Achievement                    \\
        \midrule

\rowcolor{gray!10} 0    & Trajectory Efficiency                    \\
        \midrule \rowcolor{gray!10} 0     & Barrier Avoidance                        \\
        \midrule \rowcolor{gray!10} 0     & Navigation Loop Resolution               \\
        \midrule \rowcolor{gray!30} 1     & Subtask Coordination                     \\
        \midrule \rowcolor{gray!30}1      & UI Interaction Accuracy                  \\
        \midrule \rowcolor{gray!60} 2     & Map Search Efficiency                    \\
        \midrule \rowcolor{gray!60} 2     & Wish Cart Accuracy                       \\
        \midrule \rowcolor{gray!90}3      & News and Financial Information Retrieval \\
        \bottomrule
    \end{tabular}
    \caption{Metrics for WebVoyager.}
    \label{tab:app_nnetnav_metrics}
\end{wraptable}

\end{document}